\documentclass[num-refs]{wiley-article}

\usepackage{amssymb}
\usepackage{graphicx}
\usepackage{amsmath,amssymb,amsfonts}
\usepackage{mathrsfs}

\usepackage{siunitx}

\usepackage{tikz}
\usepackage{pgflibraryarrows}
\usepackage{subcaption}
\usepackage{booktabs}
\usepackage{algorithm}
\usepackage[noend]{algcompatible}
\usepackage[noend]{algpseudocode}
\usepackage{dsfont}
\usepackage{url}
\usepackage{multirow}
\usepackage{graphicx}
\usepackage{amsmath}
\usepackage{bm}

\usepackage[misc,geometry]{ifsym} 
\newtheorem{defn}{Definition}
\papertype{Original Article}
\paperfield{Journal Section}

\title{Diversifying Agent's Behaviors in Interactive Decision Models}

\author[1]{Yinghui Pan}
\author[1]{Hanyi Zhang}
\author[2]{Yifeng Zeng}
\author[2]{Biyang Ma}
\author[2]{Jing Tang}
\author[1]{Zhong Ming}

\affil[1]{College of Computer Science and Software Engineering, Shenzhen University, Shenzhen, 510810, China}
\affil[2]{Department of Computer and Information Sciences, Northumbria University, Newcastle, NE7 7UX, UK}

\corraddress{Yifeng Zeng, Department of Computer and Information Sciences, Northumbria University, Newcastle, NE7 7UX, UK
\\ Ming Zhong, College of Computer Science and Software Engineering, Shenzhen University, Shenzhen, 510810, China}
\corremail{yifeng.zeng@northumbria.ac.uk;}

\fundinginfo{UKRI EPSRC Grant: EP/S011609/1 and NSFC: Grants: 62176225, 61772442 and 61836005}

\runningauthor{Pan et al.}

\begin{document}

\begin{frontmatter}
\maketitle

\begin{abstract}
Modelling other agents' behaviors plays an important role in  decision models for interactions among multiple agents. To optimise its own decisions, a subject agent needs to model what other agents  act simultaneously in an uncertain environment. However, modelling insufficiency occurs when the agents are competitive and the subject agent can not get full knowledge about other agents. Even when the agents are collaborative, they may not share their true behaviors due to their privacy concerns. In this article, we investigate into diversifying behaviors of other agents 
in the subject agent's decision model  prior to their interactions. Starting with prior knowledge about other agents' behaviors, we use a linear reduction technique to extract representative behavioral features from the known behaviors. We subsequently generate their new behaviors by expanding the features and propose two diversity measurements to select top-$K$ behaviors. We demonstrate the performance of the new techniques in two well-studied problem domains. This research will contribute to intelligent systems dealing with unknown unknowns in an open artificial intelligence world.      

\keywords{Interactive behaviors, Intelligent Agents, behavior Diversity}
\end{abstract}
\end{frontmatter}

\section{Introduction}
Understanding interactive behaviors facilitates the development of intelligent systems that involves interactions between either multiple agents or agents-and-humans. From the viewpoint of a subject agent, it expects to optimise its own decisions given what it observes from an environment shared by other agents who act in a similar way. However, the decision optimisation becomes  difficult when the subject agent can not get full knowledge about the other agents. This often occurs in a setting of competitive agents  where the  subject agent can not obtain sufficient knowledge about others in the decision making process.  For example, a ground force schedules its routine patrol based on the criminal activities without fully knowing what the criminals will precisely react.  Even in a collaborative agent environment, the agents may  not be willing to share their information with each other due to their privacy concerns. 

\begin{figure}[ht!]
\centering
\includegraphics[width=10.0cm]{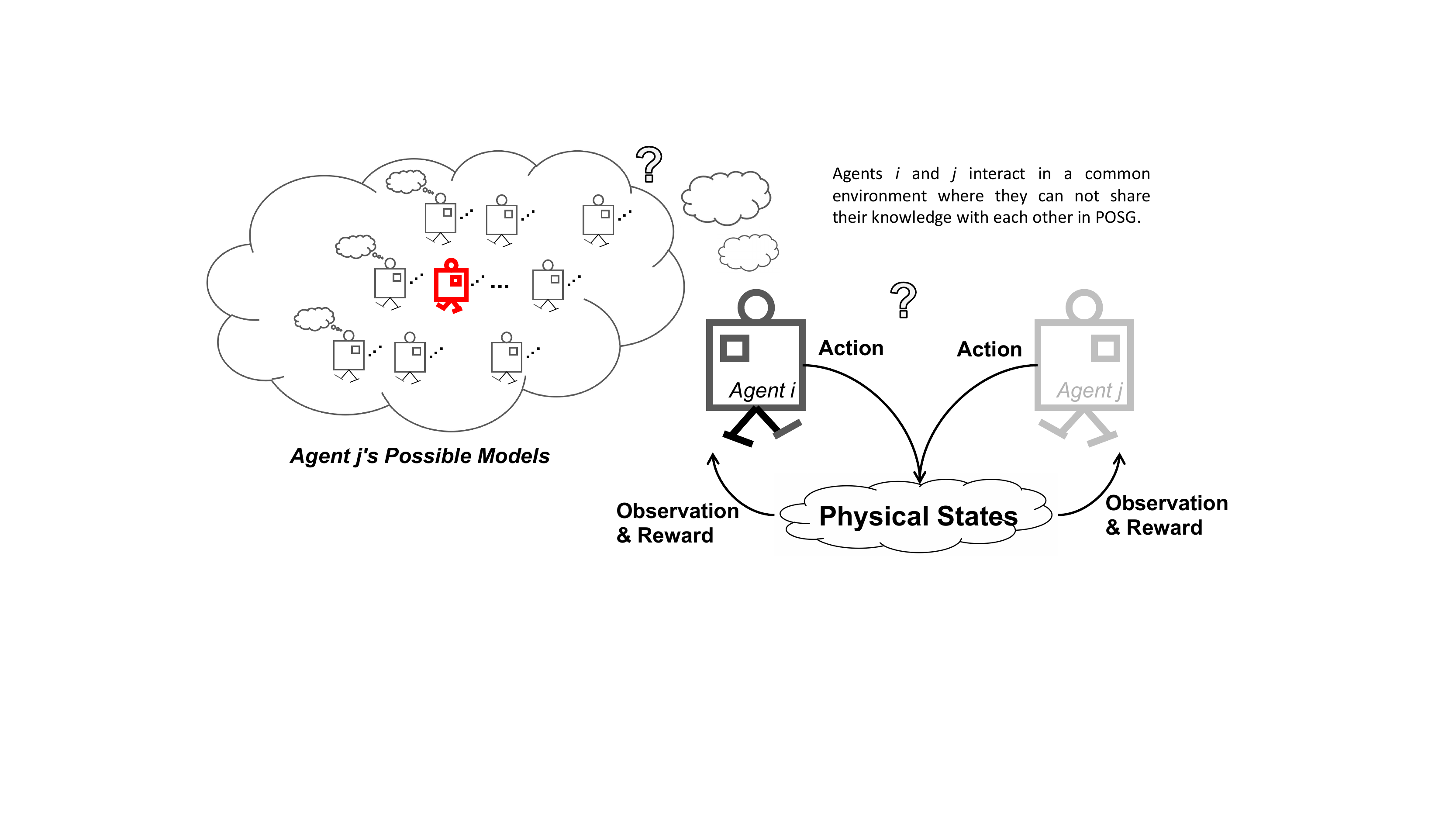}
\caption{Agent $i$ optimises its actions by solving the hypothesised models of the other agent $j$. In this case,  the true model of agent $j$~(the frame colored by red) is contained in the set of candidate models. However, it often occurs that agent $i$ does not hold the true model of agent $j$ due to the modelling insufficiency.}
\label{fig:is}
\end{figure}

Solving decision problems without sufficient prior knowledge, also termed as modelling insufficiency, has been a long-standing issue in the field of uncertainty in artificial intelligence~(UAI)~\cite{PohFH94} and opens a new challenge to deploy intelligent systems in an open AI world~\cite{AlbrechtS18}. Figure~\ref{fig:is} shows one example of interactions involving two agents~($i$ and $j$) in a partially observable stochastic game~(POSG)~\cite{Bernstein02}. Both agents act when they receive observations from the environment and are awarded according to impact of their actions on the environmental states. From the viewpoint of agent $i$, it needs to predict behaviors of the other agent $j$ which are not known prior to their interactions. Hence, agent $i$ has to hypothesise a large number of agent $j$'s possible models by solving which agent $i$ can predict $j$'s behaviors. Unfortunately, the set of candidate models may not contain the true model of agent $j$ due to the modelling insufficiency.  This leads to the challenge of optimising the subject agent $i$'s decisions under the uncertainty of other agents' behaviors. 

In this article, we consider modelling insufficiency in a well-known interactive multiagent decision model, e.g. interactive dynamic influence diagrams~(I-DIDs) and interactive partially observable Markov decision processes~(I-POMDPs), in a POSG setting~\cite{ZengD12,DoshiGD20}. As demonstrated in the previous work, I-DIDs have shown both representation and computational advantages compared to I-POMDPs~\cite{DoshiZC09,ZengD12}. Hence we develop I-DID-based solutions to the modelling insufficiency issue in this work and the solutions can also be generalized to other multiagent decision models. 
An I-DID model extends a single-agent dynamic influence diagram~\cite{Tatman90} to represent how a subject agent solves a sequential decision problem involving other agents in a common environment. In a POSG setting, both agents act simultaneously and they can not directly see what the others act and can only reason with their actions given what they receive from the environment. In addition, their observations do not fully reflect the environmental states in a deterministic manner, but with a probabilistic distribution. 

A main component of a I-DID model is a subject agent $i$'s dynamic influence diagram based on which the other agent $j$'s behaviors are modelled and embedded into the decision model.  
Modelling insufficiency occurs  where agent $j$'s behaviors are not fully represented in an agent $i$'s I-DID model. For example, if some actions of agent $j$ are not modelled, agent $i$ may not receive consistent observations in their real-time interactions. This is because agent $j$ acts according to its own optimal decisions and this true behavior is not in agent $i$'s mind when agent $i$ builds its I-DID model. The issue of inconsistent observations was also identified in a general probabilistic graphical decision model~\cite{Sondberg-JeppesenJZ13}. To the best of our knowledge, our work is the first attempt to solve the modelling insufficiency issue in I-DIDs.

Given limited amount of prior knowledge about agent $j$'s behaviors, we aim to develop a set of its new behaviors that are to be modelled in agent $i$'s I-DID models~\footnote{To facilitate the presentation, we elaborate the development using two agents~(agents $i$ and $j$) in the following discussion; however, the techniques are general for a multiagent setting.}. The new behavior set expects to contain agent $j$'s true behaviors in a large probability so as to increase agent $i$'s decision quality. Since the prior knowledge provides essential information about the other agent, we first use a linear reduction technique to elicit behavioral features from the known behaviors. The features could reflect representative patterns of what the other agent acts given its beliefs about the environment. The reduction differs from the previous work on reducing agent $j$'s behavior space since it is to find representative behavior patterns and does not aim to compress the behavior space~\cite{ZengD12,ConroyZT16,PanMTZ22}.

Subsequently, we randomize a set of new behaviors for the other agent based on the behavioral features. The randomization is conducted by sampling the action that achieves the best reward for the other agent. By doing so, we can obtain a large set of candidate behaviors for the other agent. The issue lies in selecting a set of $K$ behaviors, namely top-$K$ behaviors, that holds a good chance of including the true behaviors of the other agent. We propose two new measurements that quantify the diversity of a set of behaviors and are used to optimise the top-$K$ behavior selection. The first diversity measurement considers the difference  between specific actions upon observations at each time step while the second one contains an extra factor of differentiating general behavior patterns over time. 
Intuitively, a set of diverse behaviors have a large chance of containing the true behavior of the other agent which is not known by a subject agent prior to their interactions.  Once we have the top-$K$ behaviors, we can represent them in the subject agent's I-DID model and solve the decision model, which results in an optimal policy for the subject agent. Finally, we conduct comprehensive experiments in two well-known problem domains and show the performance of the new techniques. This work is one of very few  attempts on developing behavior diversity in I-DID solutions and will inspire more interesting work in studying agent behaviors in intelligent systems.  

We organise this article as follows. Section~\ref{sec:review} reviews the related works on modelling other agents' behaviors in interactive decision models. We present the background knowledge of I-DID models in Section~\ref{sec:idid}. In Section~\ref{sec:methods}, we propose the new methods in generating a set of new behaviors through top-$K$ behavior selection. We  demonstrate the empirical performance of our proposal in Section~\ref{sec:experiments} and conclude this work with discussions in Section~\ref{sec:summary}.

\section{Related Works}
\label{sec:review}

Research on modelling other agents has attracted growing interests in the fields of artificial intelligence, decision science and general intelligent systems~\cite{rl,Stefano20}. It mainly explores various types of modelling languages to represent decision making, behavior reasoning and learning problems in different types of environments, e.g. from a  classical model of fictitious play~\cite{brown:fp1951} to a  probabilistic deterministic finite-state automaton for modelling stochastic actions in POSG~\cite{Panella:2017}.  More sophisticated ones include recursive modelling methods that follow a nested reasoning form of what does agent $A$ think that agent $B$ thinks that agent $A$ thinks (and so on), e.g. I-POMDPs~\cite{Gmytrasiewicz05:Framework:JAIR}, and even more rigorous planning systems based on epistemic logic~\cite{Thomas:DEL}. Recently, Ma {\it et al.}~\cite{ma21} used knowledge graph to model opponents' behaviors and inferred agents' intentions accordingly. 

In particular, a series of probabilistic graphical models have been proposed to solve multiagent decision making problems. Suryadi and Gmytrasiewicz~\cite{Dicky99:ID} used influence diagrams to model other agents, but do not provide a mechanism to update the models upon a subject agent's observations. Koller and Milch~\cite{maid} proposed multiagent influence diagrams~(MAID) to compute Nash equilibrium strategies for all agents involved in the interaction while Gal and Pfeffer~\cite{nid} developed networks of influence diagrams~(NID) for recursively modelling other agents. Both MAID  and  NID  formalisms  focus  on  a  static,  single-shot  interaction.   In  contrast,  I-DIDs  offer solutions over extended time interactions,  where agents act and update their beliefs over others’ models which are themselves dynamic.

Due to unknown behaviors~(true models) of other agents, most of the research hypothesizes a large number of models for the other agents based on which a subject agent adapts its decisions even when their behaviors change over time~(sometimes is referred as a dynamic opponent)~\cite{DoshiGD20}.	This leads to a significantly increasing complexity in solving other agents' models. Hence, a large amount of research has been invested into reducing the model space of the other agents. For example, the concept of minimal mental models was used to compress candidate models of other agents so as to reduce the computational complexity due to introducing redundant potential models~\cite{PynadathM07}. The behavioral equivalence principle becomes a commonly used technique to group candidate models  that exhibit identical behaviors of other agents~\cite{Bha06, ZengDCPMC16}. Similarly,  a value equivalence was used by Conroy~\cite{Ross16} to cluster the models that have similar influence on a subject agent's expected rewards, which leads to more compressed model space. This line of work on compressing model space is also due to limited prior knowledge on building a good set of candidate behavior models. Hence, generating a good set of initial models becomes important since it may provide a good chance of containing the true models and the model compression techniques are not strictly required. 
	
Most of the current research on modelling other agents still relies on hand-crafted models according to available domain knowledge. Recently, Pan {\it et al.}~\cite{Pan0MZ021} learned models of other agents from historical data of agents' interactions, but did not provide new models from the learned models.  In this article, we aim to generate new models of other agents given limited knowledge about their behaviors and the new behaviors may not be seen in the previous interactions.

\section{Background Knowledge on I-DIDs}
\label{sec:idid}

As we will develop  methods to generate new behaviors for other agents in I-DID models, we  proceed to provide background knowledge on I-DIDs.  
By extending influence diagrams~\cite{pgm} that are commonly used to represent a single-agent decision making problem, I-DIDs are a general probabilistic graphical representation for solving interactive multiagent decision making problems under uncertainty~\cite{Doshi09:Graphical}. From the viewpoint of a subject agent, an I-DID model represents how it solves the decision problems when considering other agents' behaviors that impact their decision outcomes. It integrates multiagent game theory into individual decision making frameworks. Details about I-DIDs can be found in the previous research~\cite{ZengD12}. 

\begin{figure}[ht!]
\centering
\includegraphics[width=12cm]{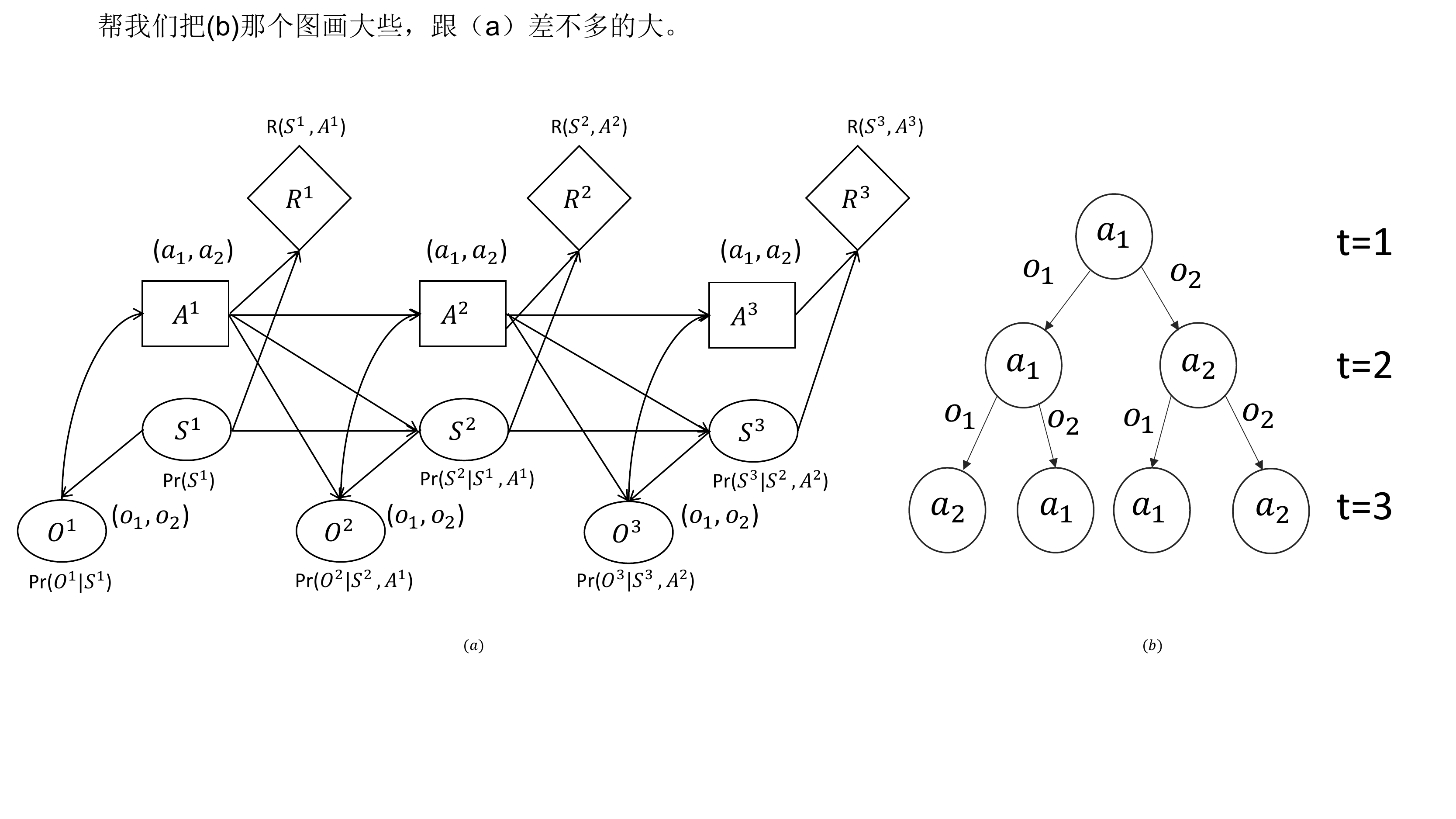}
\caption{A dynamic influence diagram and its solutions: ($a$)the dynamic influence diagram with three time steps; and ($b$) its solution is represented as a policy tree.}
\label{fig:did}
\end{figure}

Figure~\ref{fig:did}($a$) shows a dynamic influence diagram~(DID - a dynamic version of influence diagrams over time) for a single agent who plans its decisions over three time steps. In the DID model, a chance node~(denoted by a circle/oval shape) represents environmental states~($S$) and  observations~($O$) received by the agent, a decision node~(denoted by a rectangular shape) represents the agent's decisions~($A$) and a utility node~(denoted by a diamond shape)  models the agent's rewards~($R$, also called as a utility function) upon the states and decisions. The agent decision~($A$) is directly influenced by what it observes~($O$) from the environment while the environmental states~($S$) are not fully observable. To quantify the strength of variables' relations, we associate the arcs with conditional probabilities~($Pr(\cdot|\cdot)$)  while the arcs into the utility node models actual reward/utility values. For example,  $Pr(S^3|S^2, A^2)$ is a transition function from $S^2$ to $S^3$ given the effect of the action $A^2$ at time $t=2$. $Pr(O^3|S^3, A^2)$ is an observation function that models the probability of how the observations reflect the true environmental states.

Once we have the parameter specification, e.g. the transition, observation and utility functions,  in a dynamic influence diagram, we can solve the model through well-developed inference algorithms~\cite{pgm} and obtain an optimal plan for the subject agent. The plan prescribes how the agent shall act upon its observations at each time step and is often represented by a policy tree. For example, Fig.~\ref{fig:did}($b$) shows one policy tree as the optimal plan given the dynamic influence diagram in Fig.~\ref{fig:did}($a$). The agent takes action $a_1$ at the first time step $t$=1, and repeats the action if it receives observation $o_1$; otherwise, it takes $a_2$ at $t$=2. 
As a solution to a dynamic influence diagram, a policy tree represents the optimal plan according to which a subject agent behaves over times. In general, the policy tree is termed as a behavioral model - optimal decision outcomes from a decision model.  

\begin{figure}[ht!]
\centering
\includegraphics[width=10cm]{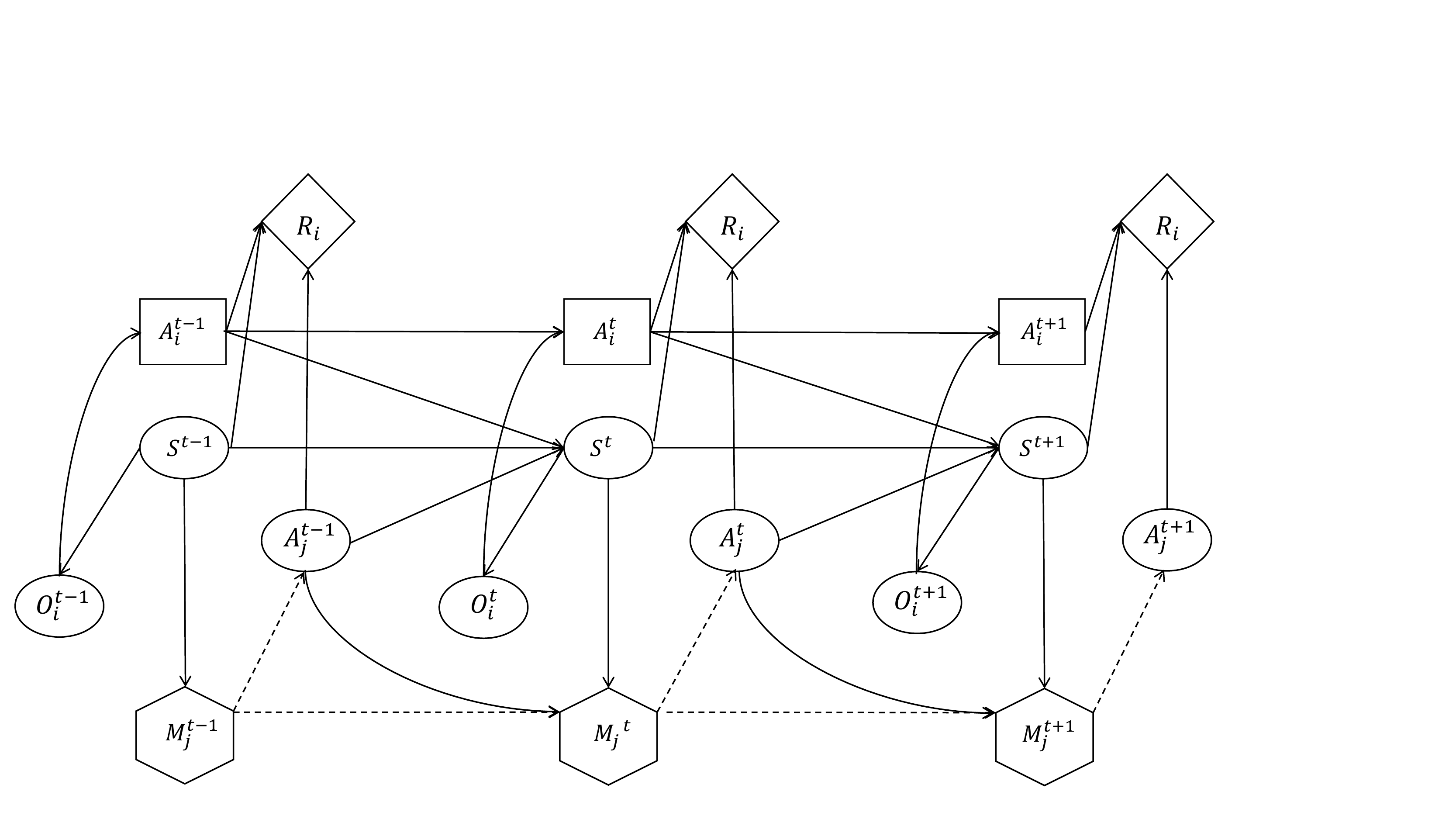}
\caption{By extending dynamic influence diagrams, the I-DID model represents how the subject agent $i$ optimises its decisions by modelling the other agent $j$'s behaviors over three time steps.}
\label{fig:idid}
\end{figure}

An I-DID model extends dynamic influence diagrams for multiple agents by introducing the model node $M_j$~(the hexagon node) in the model, as shown in Fig.~\ref{fig:idid}. The model node contains possible models of the other agent $j$ each of which is either a decision model, e.g. dynamic influence diagrams, or a behavioral model. Once the subject agent expands the model node with possible models of other agents, it can solve the I-DID model through conventional dynamic influence diagram algorithms therefore resulting in an optimal plan for the subject agent.  
As the true model is not known by the subject agent $i$, the number of candidate models contained in the model node tends to rather large. In theory, it requires an infinite number of $j$'s candidate models so that a subject agent can sufficiently predict $j$'s behaviors in order to optimise its decisions. The computational limit prevents a large model space in the model node within an I-DID model. Hence, there is a significant amount of research on dealing with the compression of the model space in the I-DID~\cite{ZengDCPMC16,Pan0MZ021}. 

Most of the current I-DID research still assumes that the true behavior is within the model node~($M_j$) and does not handle unknown behaviors of other agents that are not contained in the model node~\cite{ZengDCPMC16,Pan0MZ021}. One barrier to deal with the case of unknown behaviors is partially due to the incapability of generating a good set of candidate behaviors for other agents. The current work always stands on the pre-defined models that lead to a set of monotonic behaviors for other agents. Consequently, there is a rather large chance that the true behaviors slip from the model node in a subject agent's I-DID model. In this article, we will propose a novel model generation approach to modelling unknown behaviors in an I-DID model and focus on measure the quality of the new set of generated behaviors.

\section{Top-$K$ behavior Selection}
\label{sec:methods}

Intuitively, a large set of new behaviors have the potential of increasing the chance to include the true behaviors of other agents. However, due to the computational constraints, solving I-DID models becomes infeasible when the set is too large as shown in the previous I-DID research~\cite{ZengD12}. Hence, the set of new behaviors expect to have a good diversity while keeping a reasonable size. In addition, the new behaviors  can not be fully randomised 
since we have prior knowledge~(although it is limited) about other agents' behaviors. Accordingly, we proceed to exploit the knowledge  by eliciting representative behaviors from the known behaviors, based on which the set of new behaviors are generated and measured in terms of their diversity. 

\subsection{Eliciting Representative Behaviors}

An agent's behavior prescribes what the agent shall do given its observations in an environment. By solving a decision model, e.g. dynamic influence diagrams in Fig.~\ref{fig:did}($a$),
we can obtain the agent's optimal policy that contains a set of actions given various observations received by the agent over time. In general, the policy can be represented in a tree structure as shown in Fig.~\ref{fig:did}($b$).  Each branch of the policy tree is a behavior sequence  specifying the agent's optimal action given a possible observation at each time step.  We formally define a behavior sequence below.

\begin{defn}[Behavior Sequence]
 For agent $j$, a behavior sequence, $h_T^j=(a_1, o_2, a_2, \cdots, o_{T-1}, a_{T})$ where $a_t$~($\in A$) and $o_t$~($\in \Omega$),  is a set of alternating actions and observations over its planning horizon $T$. 

\label{defn:bs}
\end{defn}

Subsequently, we can define a policy tree that is composed of a set of behavior sequences. 

\begin{defn}[Policy Tree]
A policy tree for agent $j$ is a set of behavior sequences, ${H}_T^j$=$\bigcup h_T^j$, that are organized as a tree structure with the depth $T$ where actions are in the nodes while observations are attached to the branches in the tree.
\label{defn:pt}
\end{defn}

Following a policy tree, an agent executes a behavior sequence when it receives a particular observation  from an environment at each time step.  In the policy tree example in Fig.~\ref{fig:did}($b$), the agent takes the action $a_1$ at the first time step and will execute the actions $a_2$ and $a_1$ given its observations $o_2$ and $o_1$ respectively at the second and third time steps. 

A policy tree represents the agent's behavior that can be obtained by solving its decision model. Variations of parameters, e.g. probability distributions in a dynamic influence diagram, in the agent's decision model could lead to different behaviors. From the viewpoint of a subject agent $i$, what does matter is the behaviors exhibited by the other agent $j$, not how agent $j$ optimises its behaviors through a decision model.  Hence, in this article, we will focus on $j$'s behavior representation and assume that  agent $i$ knows a set of $M$ behaviors for the other agent $j$. The question remains on generating a new set of behaviors~($\geq M$ including the known behaviors) for agent $j$ based on the known $M$ behaviors. The set of new behaviors are expected to include potentially true behaviors of the other agent $j$.

The known $M$ behaviors represent basic types of how agent $j$ behaves according to agent $i$'s prior knowledge about agent $j$. They can serve as salient features to expand new  behaviors of agent $j$. We proceed to extract a set of behavior sequences from the known agent $j$'s behaviors based on which its new behaviors are to be generated.  The behavior sequences are representative of agent $j$ behaviors that are known to agent $i$.

\begin{figure}
  \centering
  \includegraphics[width=3.50in]{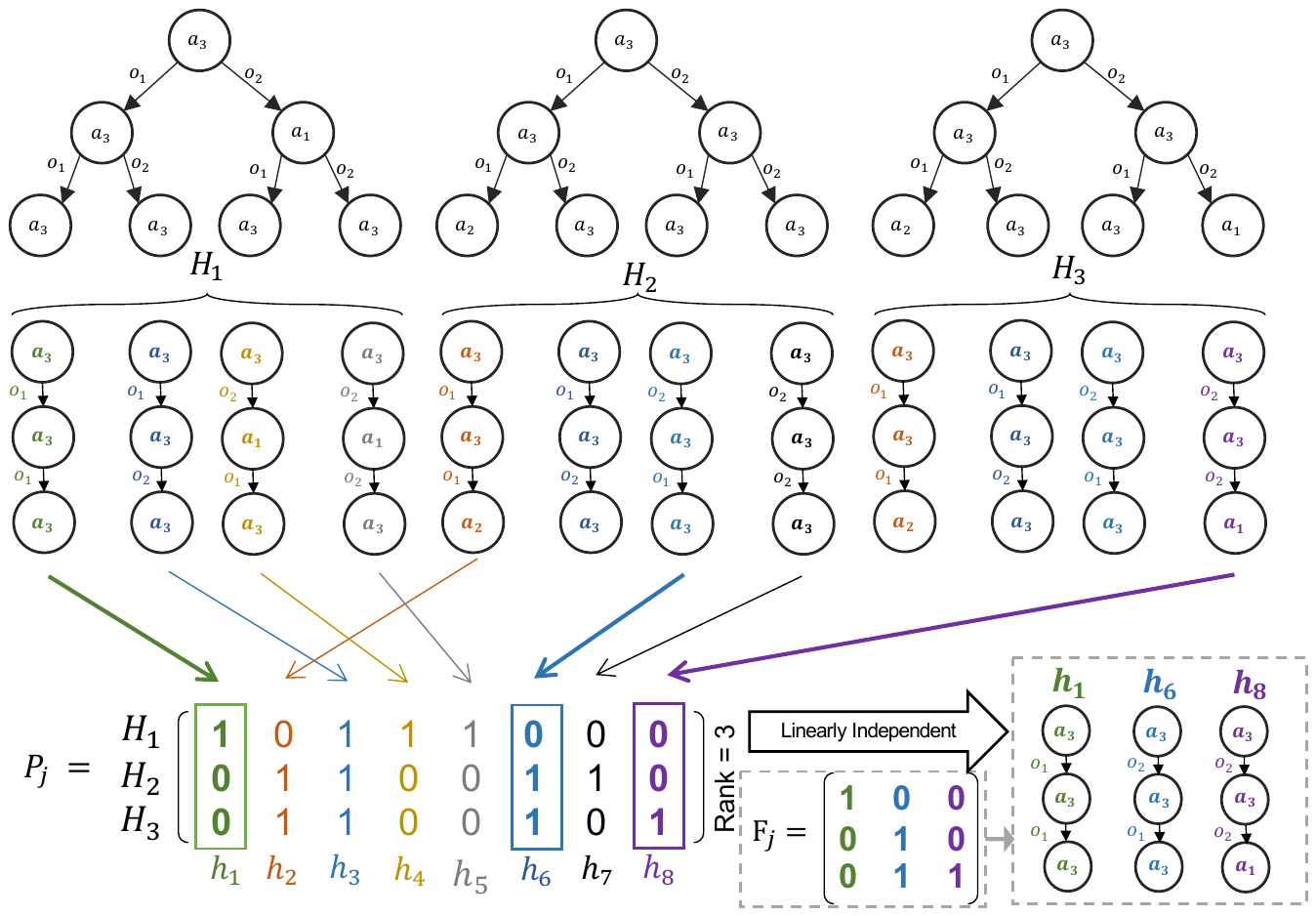}
  \caption{A set of behavior sequences, $F_j$=($h_1$, $h_6$, $h_8$), are selected from known behavior matrix $P_j$ for agent $j$.}
  \label{fig:bf}
\end{figure}

 Instead of randomly selecting $m$ behavior sequences, $F_j$=$\{h_1, \cdots, h_m\}$, from  $M$ behaviors, $\mathcal{H}$=$\{ {H}_1, \cdots,  {H}_M \}$~\footnote{We use the subscript of $H$ according to the context. Here it numbers the policy tree in the set while it may refer to the length of a policy tree as defined in Def.~\ref{defn:pt}. It shall be self-evident in the context.}, we use a linear reduction method to extract the sequences that have most sufficient features from the known behaviors. We compose a behavior matrix, namely $P$, where each row is a policy tree  ${H}_m$ and each column is one behavior sequence seen in the policy trees. Since some behavior sequences could appear in different policy trees, the column dimension is smaller than $M \times |\Omega|^{T-1}$, but much larger than the row dimension $M$ particularly for a large planning horizon $T$.  The matrix element $P({H}_i, h_j)$ is 1 if the sequence $h_j$ appears in the behavior ${H}_i$; otherwise, $P({H}_i, h_j)$ = 0. Hence, the matrix $P$ may contain some linearly dependent behavior sequences so that its columns could be reduced into a set of representative sequences $F$. To extract the linearly independent sequences $F$, we use a $Gaussian$ elimination method to find the pivot columns in the large matrix $P$. The extraction process can be conducted in a polynomial time~\cite{JMLR16}. In principle, we find one pivot matrix $F_j$~(corresponding to the linearly independent sequences $F$~\footnote{To simplify the presentation, we do not differentiate the two terms between a pivot {\em matrix} and {\em linearly independent sequences}.}) and the other matrix $U_j$ to ensure: $P_j$=$F_j \times U_j$.
 
In Fig.~\ref{fig:bf}, we elaborate the extraction of behavior sequences from three behavior models $\mathcal{H}=\{{H}_1, {H}_2, {H}_3 \}$. The behavior matrix $P_j$ has the rank of three so that we could obtain the pivot matrix  $F_j$=
\begin{equation}
\begin{bmatrix}
1 & 0 & 0\\
0 & 1 & 0\\
0 & 1 & 1
\end{bmatrix}
\end{equation}
that returns three linearly independent sequences, e.g. $F_j$=$(h_1, h_6, h_8)$. The selected sequences provide sufficient features, also called as behavioral features, that represent the three known behavior models.

We shall note that selecting the matrix-based behavior sequences is different from the research on compressing model or behavioral space~\cite{PynadathM07,ZengD12}. The previous compression method focuses on reducing the entire behavior space while maintaining a complete policy tree. Our work is to select a set of representative behavior sequences that are often a partial policy tree. The selected sequences provide important behavioral features in the expansion of new behaviors for other agents.  

\subsection{Measuring Behavior Diversity}

Given the representative behavior sequences $F_j$, we aim to generate  new policy trees that contain the sequences corresponding to a set of branches,  and select the top-$K$ policy trees  by measuring their diversity. Algorithm~\ref{algorithm:topk} presents the procedure of top-$K$ policy tree generation. 

\begin{algorithm}[ht]
\caption{Generating Top-$K$ Policy Trees $\mathcal{H}_K$}
\label{algorithm:topk}
\begin{algorithmic}[1]
\Function{TopK}{$F_j$, $\mathcal{H}$=$\{$ $H_1$, $\cdots$, $H_M$ $\}$ and DID of Agent $j$}
\State Get a set of behavior sequences $F_j$ from $\mathcal{H}$
\State Convert agent $j$'s DID into dynamic Bayesian networks $B_j$
\State Instantiate $B_j$ with $F_j$
\State Initiate $\mathcal{H}_K$=$\{$ $H_1$, $\cdots$, $H_M$ $\}$
\Repeat
\State Sample a full policy tree $H_T$ from the $B_j$ 
\State $\mathcal{H}_K$=$\mathcal{H}_K$ $\bigcup$ $H_T$ 
\State Recompute the set diversity $Div(\mathcal{H}_K)$
\Until {$Div(\mathcal{H}_K)$ does not change}
\State \Return $\mathcal{H}_K$
\EndFunction
\end{algorithmic}
\end{algorithm}

To sample a full policy tree, we first convert agent $j$'s dynamic influence diagram into its counterpart of Bayesian networks $B_j$ and instantiate the networks using the known behavior sequences $F_j$~(line 3-4). We convert decision nodes $A$ into chance nodes where we can instantiate their states with the actions $a$ from the behavior sequences $F_j$. The utility nodes are converted into chance nodes where the utility values are normalized into probabilities in the nodes. 
For chance nodes $O$, we instantiate their states with the observations $o$ from $F_j$. 

We randomise a probability distribution in the initial beliefs $S^1$ in $B_j$ and calculate the probability distributions for the chance nodes $O^t$ and $A^t$ at each time step. Given the probability $Pr(O^t|S^1, F_j)$, we sample a possible observation $o_t$ to be added into a new tree $H_T$. To decide the action $a_t$ given $o_t$, we choose the best one that results in the largest utility value $R(S^t,A^t)$. By sampling the actions and observations over the planning horizon $T$, we can compose a set of behavior sequences therefore composing a new policy tree $H_T$~(line 7). In other words, we complete a full policy tree by filling in the rest of its branches based on the known sequences. We repeat the adding of new policy trees until the diversity of the new set of $K$ policy trees does not increase~(line 8-9). We may terminate the process once a big $K$ is reached. This is to ensure that the I-DID models with the top-$K$ behaviors can be solved within a computational limit. 

What remains is to compute the diversity of the behavior set $Div(\mathcal{H}_K)$. We need to measure how different the $K$ behavior trees would be in terms of their actions given specific observations over $T$ time steps. We propose two diversity measurements for this purpose. The first measurement of diversity considers difference among behavior sequences in a vertical manner: the sequences or paths~(with one specific observation at one time step) in the tree are examined separately along the depth. It, called as MDP~(measurement of diversity over paths), mainly measures the diversity of sequences in the tree.  The second diversity measurement extends MDP with the extra consideration of behaviors in a horizontal way: the sequences~(with all possible observations at one time step) are compared along the width. It, also named as MDF~(measurement of diversity with frames), measures the frames of a policy tree on the top of MDP. 

In the first measurement, we retrieve all the different behavior sequences  from the policy trees $\mathcal{H}_K$. For every sequence $h_T$, we aggregate all sub-sequences $h_t$ with the length $t$~($\in$ [$1,T$]). Since early actions of agent $j$ have immediate impacts on agents' interactions,  the short sequences contribute more into the diversity. Hence, we weight the sequence $h_t$ using the factor of $\frac{1}{|\Omega|^{t-1}}$. Formally, we define the MDP diversity of $K$ policy trees in Eq.~\ref{eqn:div1}.

\begin{equation}
    Div(\mathcal{H}_K)_{MDP} = \sum_{t=1}^T \frac{Diff(h_t)}{|\Omega_j|^{t-1}}
    \label{eqn:div1}
\end{equation}
where $Diff(h_t)$ is the number of different sequences $h_t$ in $\mathcal{H}_K$ and |$\Omega_j$| is the number of agent $j$'s observations. 

The first measurement considers a single action at one time step~(within individual sequences) and may lose a general picture of what agent $j$ behaves given different observations at one time step. To capture the frame of general behaviors, we add the diversity of sub-trees~($H_t$ as a part of each policy tree in $\mathcal{H}_K$) of different depths into MDP. This leads to the second diversity measurement MDF in Eq.~\ref{eqn:div2}.

\begin{equation}
    Div(\mathcal{H}_K)_{MDF} = \sum_{t=1}^T \frac{Diff(h_t)+Diff(H_t)}{|\Omega_j|^{t-1}} 
    \label{eqn:div2}
\end{equation}
where $Diff(h_t)$ and $Diff(H_t)$ are the numbers of different sequences $h_t$ and sub-trees~(frames) $H_t$ respectively in $\mathcal{H}_K$, and |$\Omega_j$| is the number of agent $j$'s observations.

\begin{figure}[ht!]
\centering
\includegraphics[width=10.0cm]{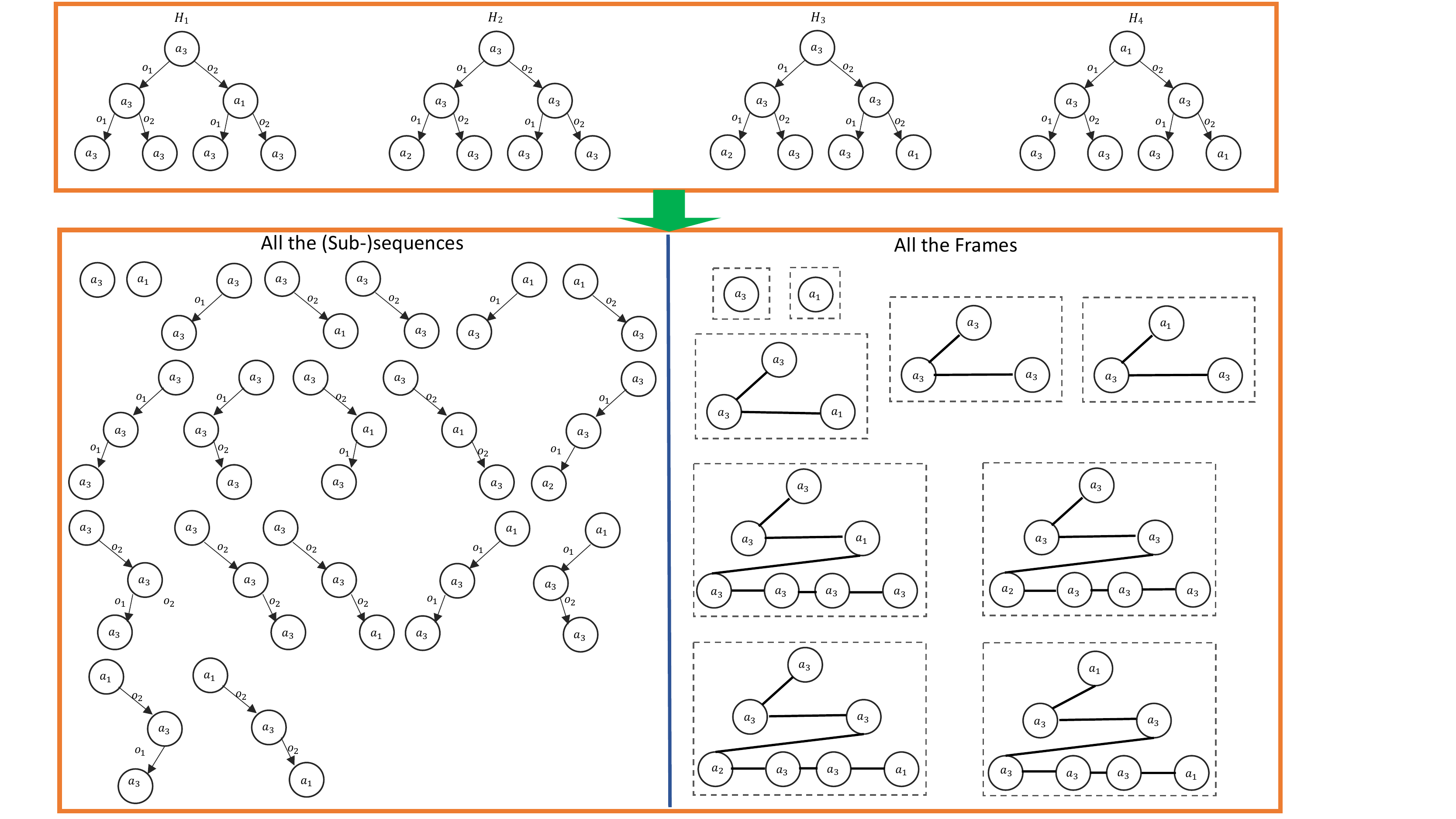}
\caption{By aggregating all the (sub)-sequences and frames from the known policy trees~$\mathcal{H}$=$\{H_1, \cdots, H_4\}$, the two diversity measurements~(MDP and MDF) can be calculated.}
\label{fig:md}
\end{figure}

In Fig.~\ref{fig:md}, we elaborate the two diversity measurements~(MDP and MDF) in one specific example. Given four behavior trees, e.g. $\mathcal{H}$=$\{H_1, \cdots, H_4\}$, we could obtain 2 different sequences with the length of one time step, 5 different sequences with the length of two time steps and 12 different sequences with the length of three time steps, as shown in the left panel in Fig.~\ref{fig:md}. In contrast, there exist: 2 different frames with one time step, 3 different frames with two time steps and 4 different frames with three time steps in the right panel. Following the diversity calculation in Eqs.~\ref{eqn:div1} and~\ref{eqn:div2}, we can get the diversity values for the four policy trees using the two measurements, $Div(\mathcal{H}_K)_{MDP}$=7.5 and $Div(\mathcal{H}_K)_{MDF}$=12.0, respectively.

As shown in Eqs.~\ref{eqn:div1} and~\ref{eqn:div2}, MDF has a larger diversity value than MDP since it includes more factors in the computation and could differentiate  more behaviors. We will investigate how this extra information will add value into decision quality for a subject agent  in the experiments. In terms of time complexity, MDP needs to compare all possible behavior sequences~(with length from 1 to $T$) for at most $N$ potential trees~(generated in the sampling process) resulting in $\frac{NT(1+T)}{2}$ comparison operations while MDF requires extra $NT$ operations in the frame comparison. Since the comparison is merely to check the actional equivalence, there is no significant difference  between the time complexities of the two diversity measurements. 

Once we have the diversity measurements, we calculate the set diversity~($Div(\mathcal{H}_K)$ in line 9-10) using either $Div(\mathcal{H}_K)_{MDP}$ or $Div(\mathcal{H}_K)_{MDF}$ in the the top-$K$ behavior selection in Alg.~\ref{algorithm:topk}. Notice that we decide whether a new policy tree is to be included in the top-$K$ set while generating one from the sampling process~(line 7). We do not generate all the policy trees and select $K$ from them. As  the number of possible new trees could be infinite, it is still an open issue about the boundary of agents' $sound$ behaviors given prior knowledge. 

\section{Experimental Study}
\label{sec:experiments}
We implemented the proposed approach of generating new behaviors for other agents by first eliciting linearly independent behavior sequences and then selecting the top-$K$ behaviors through the two behavior diversity measurements. We integrated this new approach into the I-DID solutions and investigated its performance in an empirical way. We used two well-known problem domains in our experiments. One is multiagent tiger problem while the other is multiagent unmanned aerial vehicle~(UAV) problem~\cite{ZengD12}. All the implementations and tests were conducted in Window 10 with the setting of CPU~($11^{th}$ Gen Intel Core i7-6700 @ 3.40GHz 4-core) and  24GB RAM.

For each problem domain involving two agents, we built an I-DID model for the subject agent $i$ who hypothesises a number of behavioral models of the other agent $j$ in the model node $M_j$ of Fig.~\ref{fig:idid}. 
As the new approach provides a new way of solving I-DID models by providing more diverse behaviors to the other agent $j$, we compare three I-DID algorithms in the experiments. One is the state-of-art I-DID algorithm~(IDID) that expands the model node only using the known models $M$.
However, it assumes that the true behaviors of agent $j$ are in the model node. The other two algorithms, namely IDID-MDP and IDID-MDF, use the diversity measurements of MDP and MDF  to select top-$K$ behaviors respectively for the other agent, and expand the model node in the I-DID models.  The three algorithms~(IDID, IDID-MDP and IDID-MDF) adopt the same exact algorithm to solve the I-DID models~\cite{DoshiZC09} once the model nodes are expanded with different candidate models of the other agent $j$.

We evaluate the algorithms in terms of the average rewards that agent $i$ receives when it interacts with agent $j$. We randomly select one behavioral model of agent $j$ as its true model that is either from the $M$ known models or a randomised model from $K$ models~(generated by either the IDID-MDP or IDID-MDF algorithm). Subsequently, agent $i$ executes its I-DID optimal policies while agent $j$ follows the selected behaviors in their interactions. For every interaction, agent $i$ receives actual rewards given the outcomes of their actions for each time step and accumulates the rewards over the entire planning horizon $T$. We let both agents interact for 50 rounds and compute the average rewards for agent $i$ accordingly. In addition, we investigate the impact of the diversity values for the two diversity measurements  in the experiments. 

\subsection{Multiagent Tiger Problems}

 \begin{figure}[!htb]
  \centering
  \includegraphics[width=3.00in]{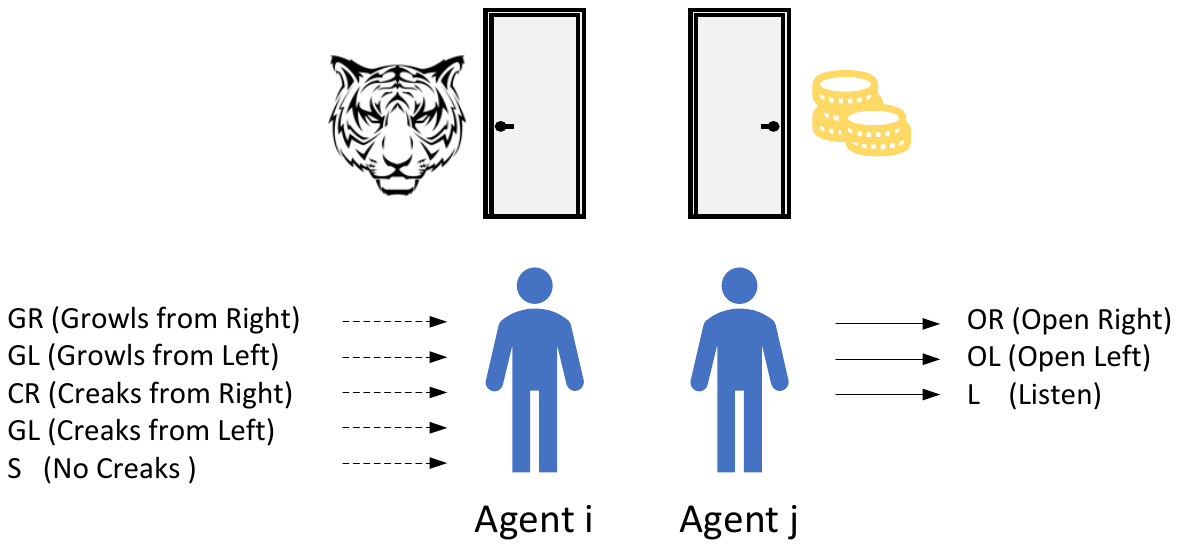}
  \caption{ A two-agent tiger problem where agent $i$ models agent $j$'s behaviors in order to optimise its own decisions over times. The problem specification follows: |$S$|=2,|$A_i$|=|$A_j$|=3,|$\Omega_i$|=6, and |$\Omega_j$|=2.}
  \label{fig:tiger}
\end{figure}

A multi-agent tiger problem is well studied in multi-agent planning research and has become a benchmark for evaluating agent planning models~\cite{Bernstein02,DoshiGD20}. We consider the two-agent version of this problem in Fig.~\ref{fig:tiger}. Both agent $i$ and agent $j$ need to decide either Open the Right/Left-hand side of door~(OR or OL) or Listen~(L) when they are uncertain of a tiger's location~(behind a door). If both open the door behind which  a pot of gold exists, they share the gold; otherwise, they will be eaten by the tiger if only one of them faces the tiger. Their decisions are based on what they can observe, e.g. tiger's growls or creaks from either door.  From the viewpoint of agent $i$, it needs to predict what agent $j$ does simultaneously in order to optimise its own decisions. We build an I-DID model for agent $i$ and vary the agent $j$'s model space $M_j$ in the I-DID model. 

\begin{figure}[!ht]
\centering
\includegraphics[width=6cm]{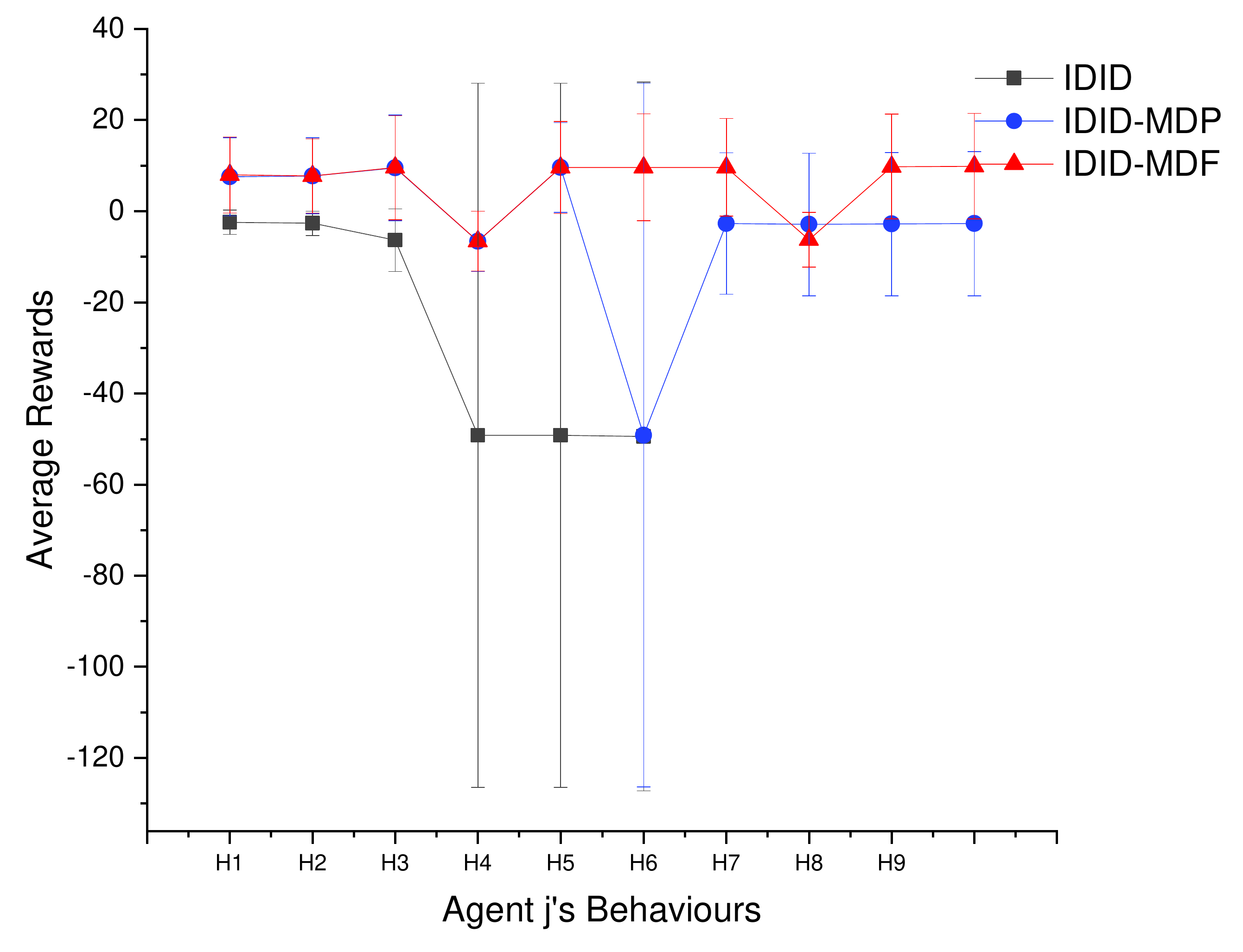}
\includegraphics[width=6cm]{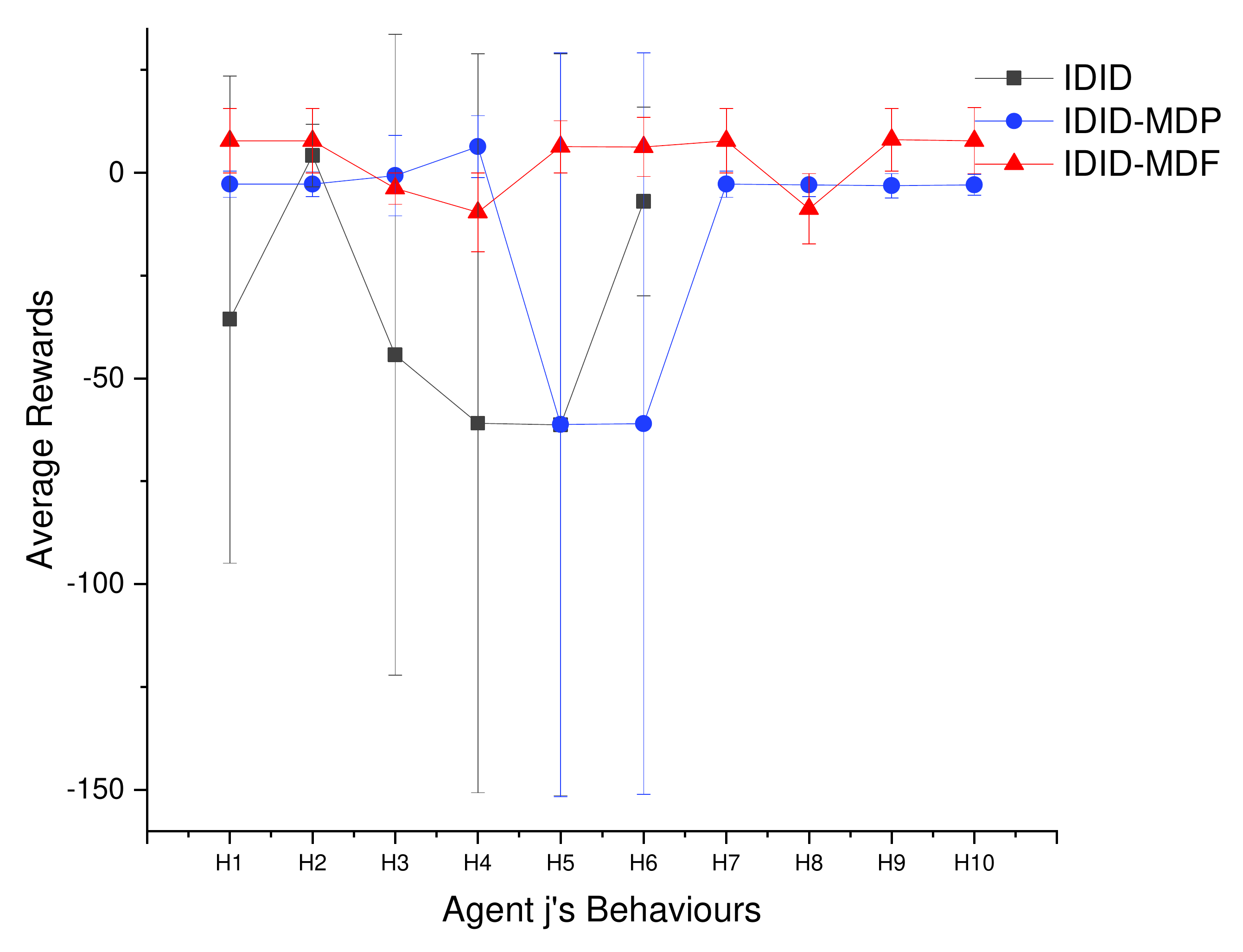}
\hspace{3cm}($a$)$T$=3\hspace{5cm}($b$)$T$=4
\caption{Average rewards received by the subject agent $i$ when it uses $j$'s behaviors models in the I-DID models of the planning horizons ($a$) $T$=3 and ($b$) $T$=4.}
\label{fig:offline}
\end{figure}

Figure~\ref{fig:offline} shows the average rewards received by agent $i$  when it runs the I-DID models with different planning horizons~($T$=3 and 4). For both the models, we have six initial models~($M$=6) for agent $j$. However, the model with $T$=4 selects four new models~($K$=4) when the model with $T$=3 adds only three new models~($K$=3) using both MDF and MDP measurements in the top-$K$ model selection. The selection is reasonable since the model with a large planning horizon often has more different behaviors. In almost all the cases, the IDID-MDF algorithm achieves better performance~(when agent $i$ receives larger rewards) than the other two algorithms. In addition, we observe that IDID-MDF has better reliability than IDID-MDP in terms of small variances~(the vertical candle bars). This is partially attributed to the merit that the MDF measurement considers a general behavioral pattern when selecting the models.

\begin{figure}[!ht]
\centering
\includegraphics[width=6cm]{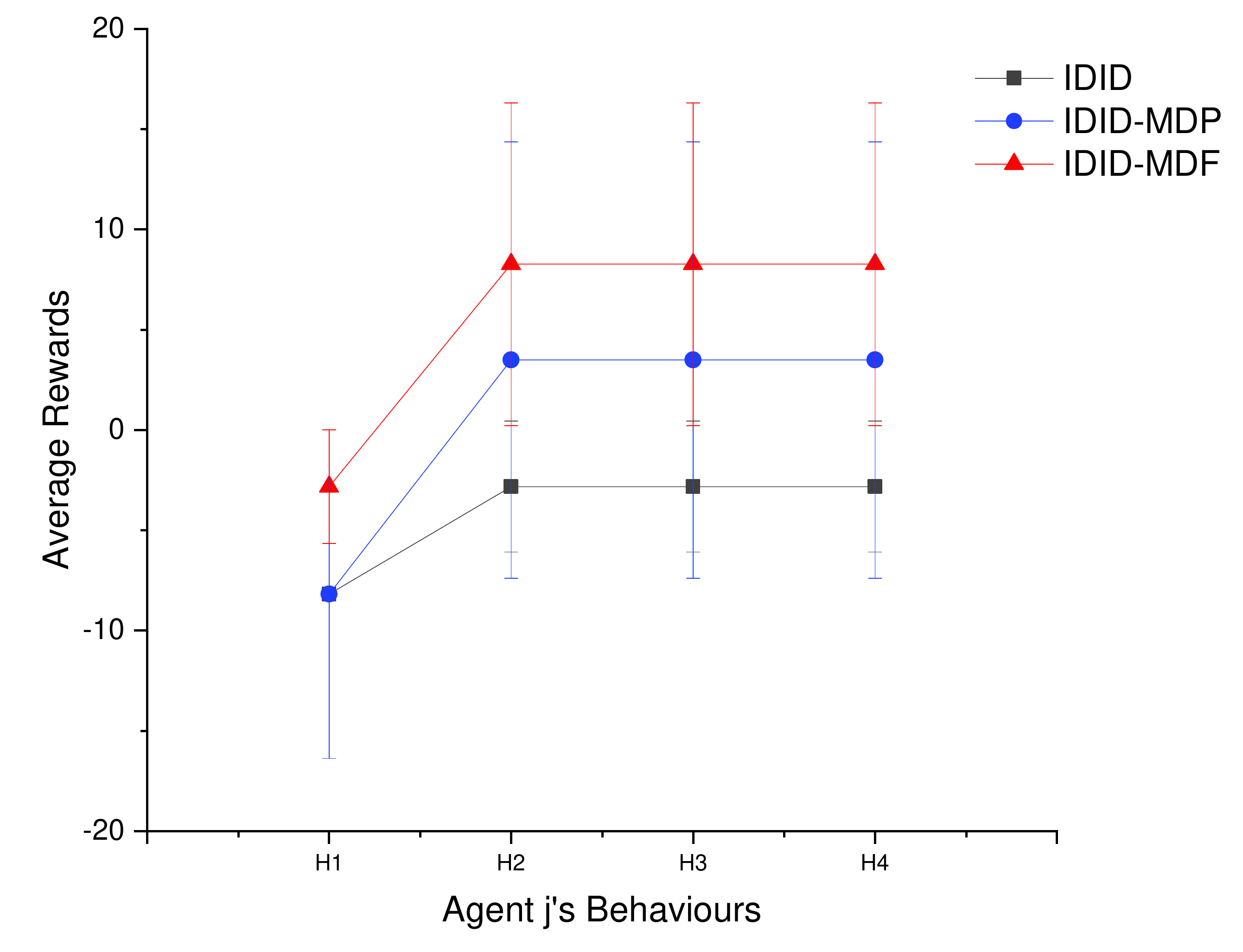}
\includegraphics[width=6cm]{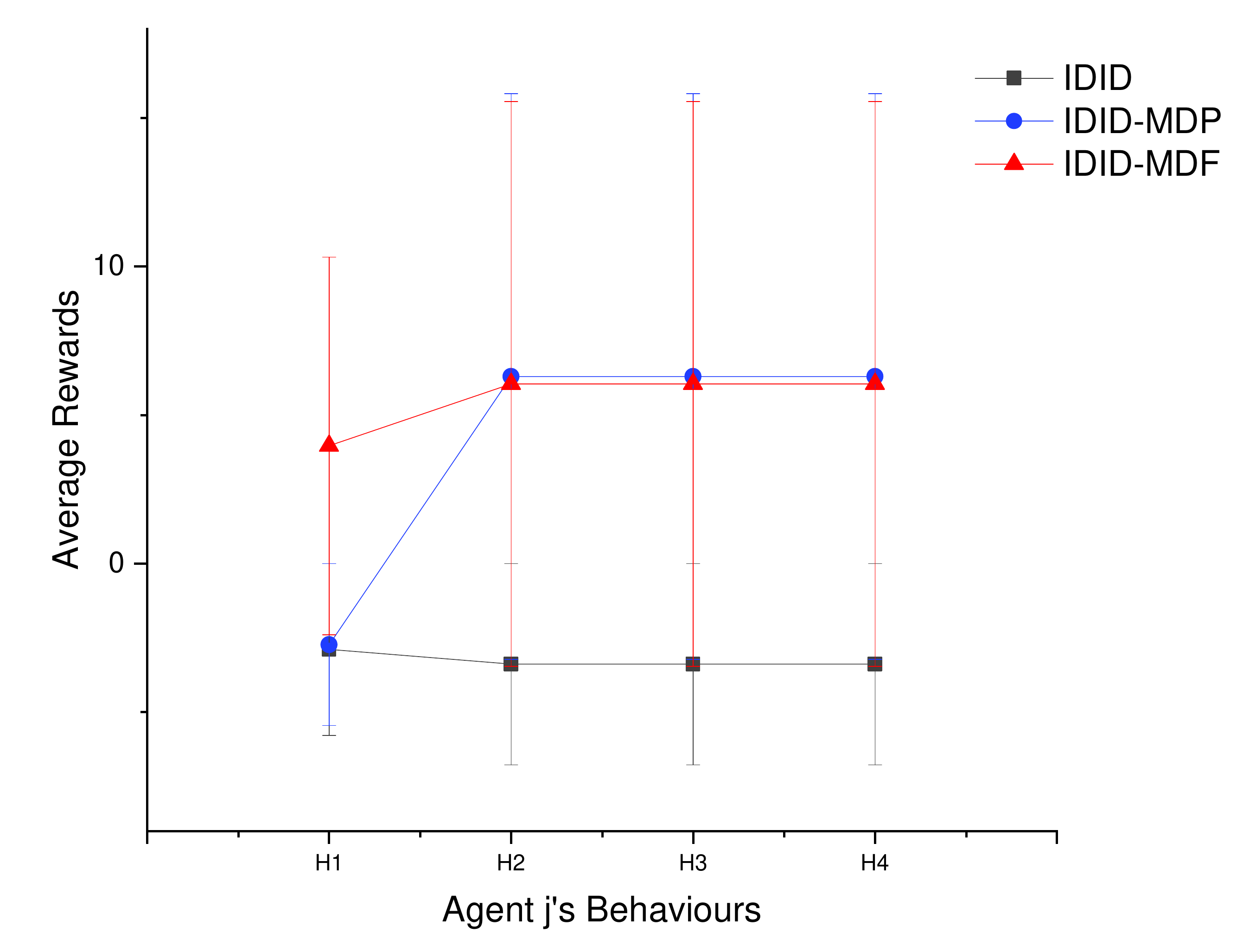}
\hspace{3cm}($a$)$T$=3\hspace{5cm}($b$)$T$=4
\caption{Average rewards received by the subject agent $i$ when the true model of agent $j$ is randomly selected from one of the candidate models within the mode node in the I-DID models with ($a$)~$T$=3 and ($b$)~$T$=4.}
\label{fig:origin}
\end{figure}

Subsequently, we run the I-DID models with $K$=4 given four initial~($M$=4) models of agents. Hence, the model size in the models are equal to eight. In the first set of experiment, we have the true model $j$ selected from the eight models~(as the experiments in Fig.~\ref{fig:offline}). Fig.~\ref{fig:origin} repeats the similar performance pattern for four models randomly selected in the experiments. 

\begin{figure}[!ht]
\centering
\includegraphics[width=6cm]{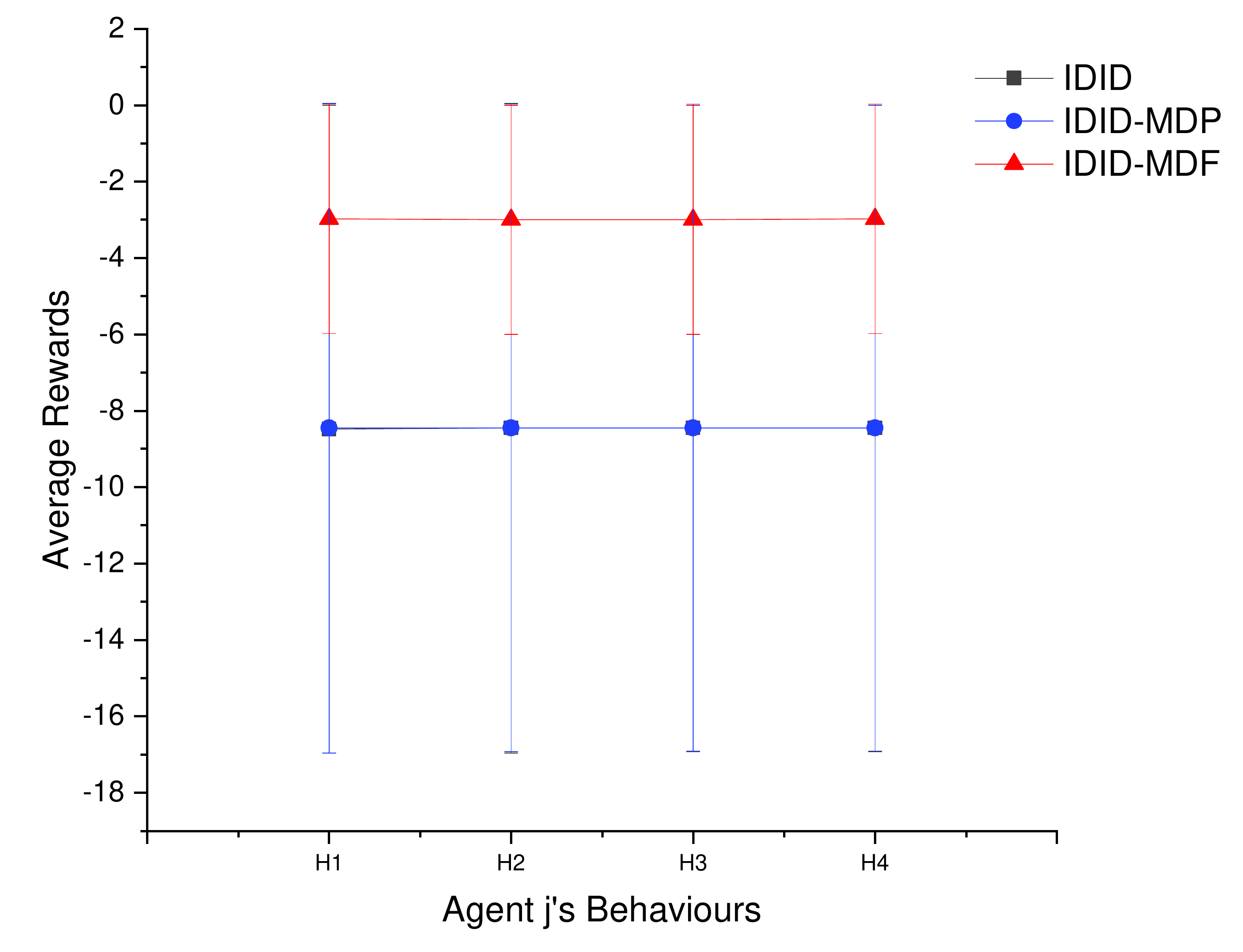}
\includegraphics[width=6cm]{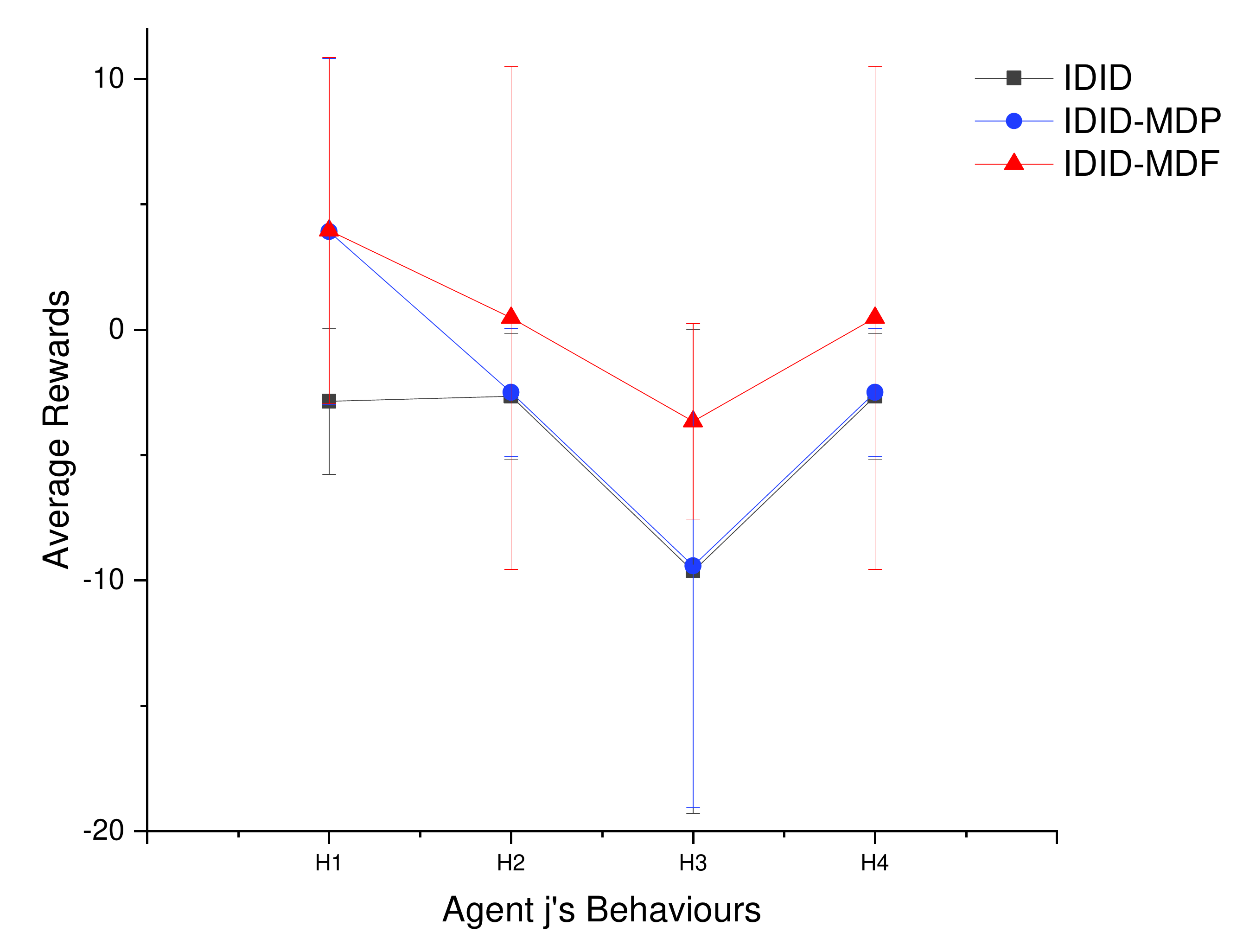}
\hspace{3cm}($a$)$T$=3\hspace{5cm}($b$)$T$=4
\caption{Average rewards received by the subject agent $i$ when the true model of agent $j$ is randomly generated~(not being identical to any model generated by IDID-MDP or IDID-MDF) in the I-DID models ($a$)~$T$=3 and ($b$)~$T$=4.}
\label{fig:random}
\end{figure}

In contrast, in the second set of experiments, we randomly generate a true model for agent $j$~(based on the initial models). The model is not identical to any model in the model node expanded by the three algorithms. Hence agent $i$ interacts with agent $j$ whose behaviors are not completely in agent $i$'s model space. 
This is to test the model capability in dealing with unknown behaviors of agent $j$. Fig.~\ref{fig:random} demonstrates that the IDID-MDF algorithm performs better than both I-DID and IDID-MDP. This is benefited from the fact that IDID-MDF includes more diverse behaviors so that it can capture some random models~(could potentially be true models) in a good manner.  

Finally, we empirically investigate the relations between the model diversity and the average rewards. For each I-DID model with the same planning horizon~(either $T$=3 or $T$=4), we run various settings of $M$ and $K$, and calculate the diversity of the resulting candidate models in the I-DID models as well as the average rewards received by agent $i$ in the interactions. 

\begin{figure}[!ht]
\centering
\includegraphics[width=7cm]{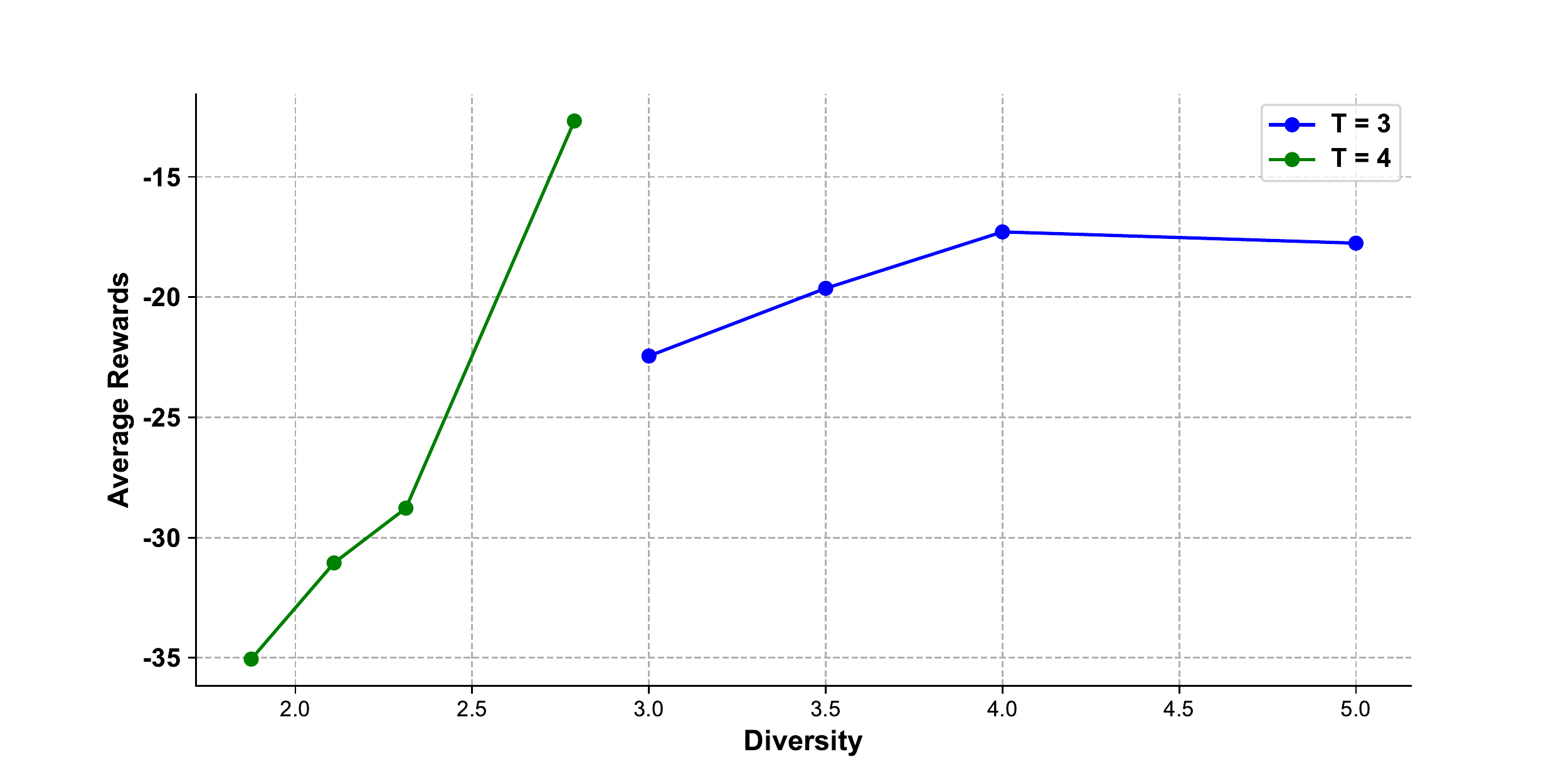}
\includegraphics[width=7cm]{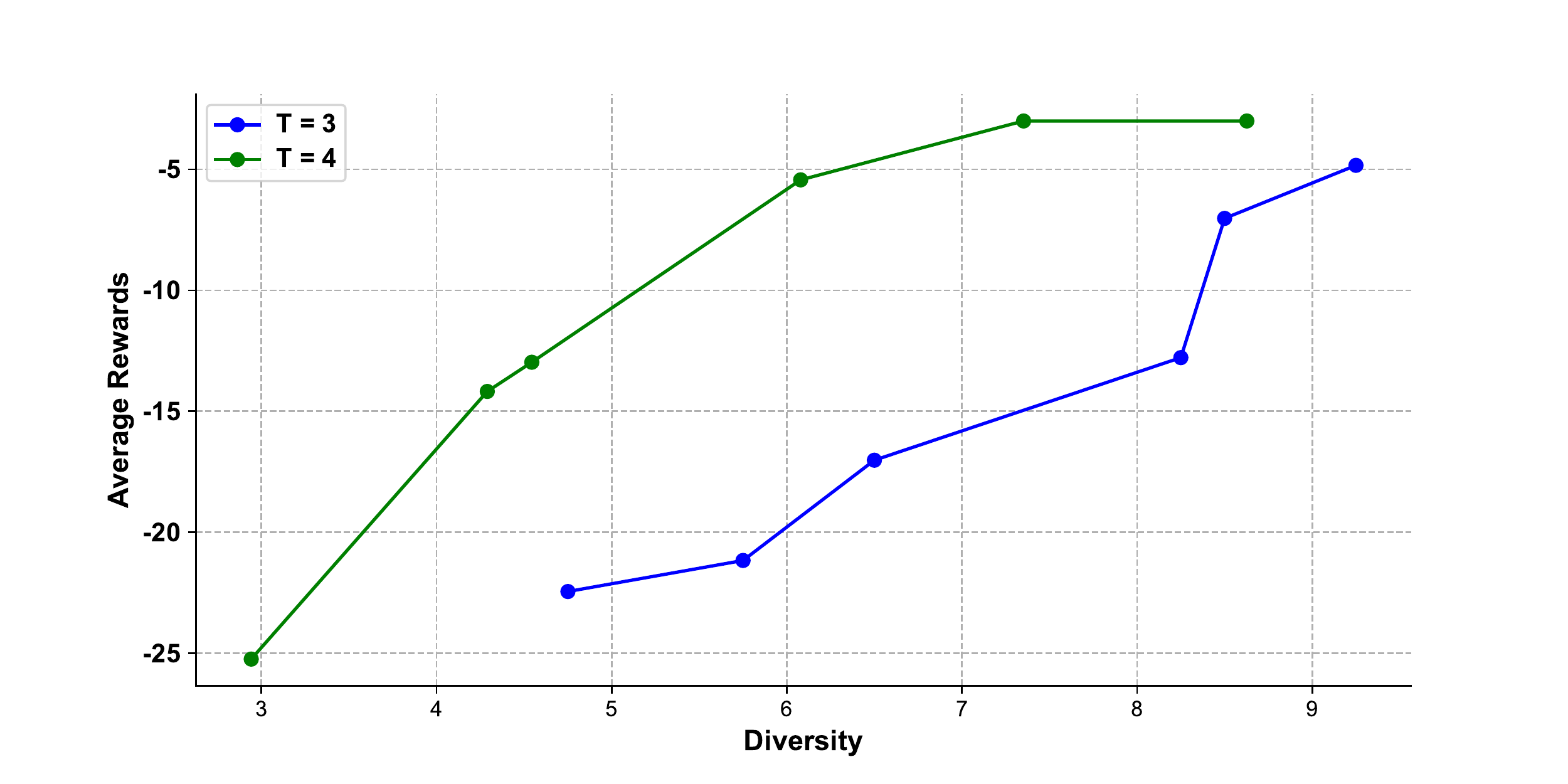}
\hspace{4cm}($a$) MDP Performance \hspace{4cm}($b$) MDF Performance
\caption{Relations between the diversity values~[of ($a$) MDP and ($b$) MDF] and the average rewards received by agent $i$ using the I-DID models with $T$=3 and $T$=4.}
\label{fig:diversityTiger}
\end{figure}

Figure~\ref{fig:diversityTiger} shows positive correlations between the diversity values and the average rewards. The MDF values are generally larger than the MDP values as indicated in Eq.~\ref{eqn:div1} and~\ref{eqn:div2}. Agent $i$ obtains better rewards when the I-DID models have larger diversity values using both the MDP and MDF measurements.  This verifies our motivation of improving agents' decision quality through diversifying model selection.

\subsection{Multiagent UAV Problems}

\begin{figure}
  \centering
  \includegraphics[width=2.00in]{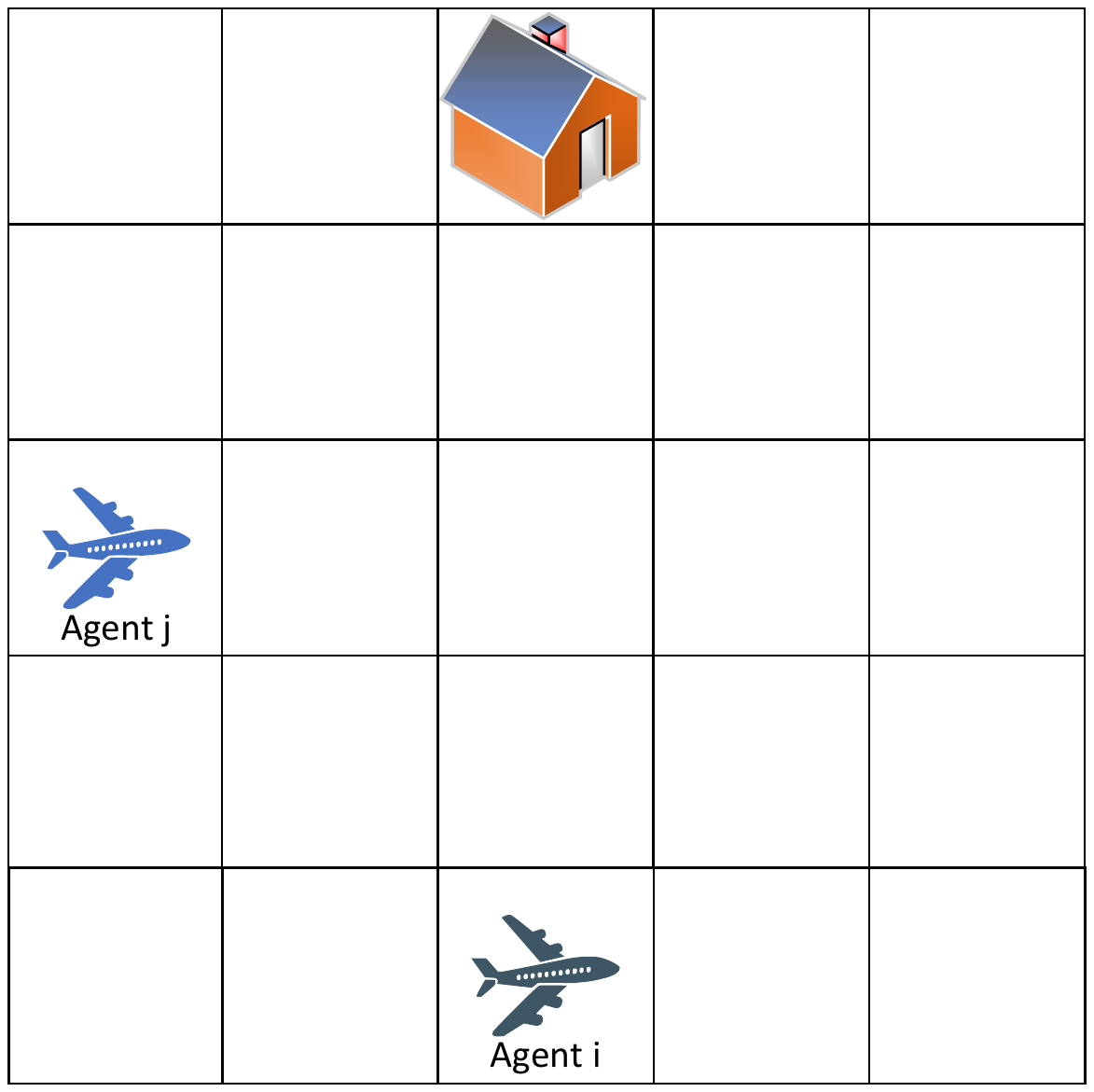}
  \caption{ A two-agent UAV problem where agent $i$ intends to capture agent $j$ before agent $j$ reaches the safe house.  The problem has the specification of |$S_i$|=81, |$S_j$|=25, |$A_i$|=|$A_j$|=5, and |$\Omega_i$|=|$\Omega_j$|=4.}
  \label{fig:uav}
\end{figure}

The multi-agent UAV problem is the largest problem domain in testing I-DID algorithms~\cite{ZengD12,DoshiGD20}. As shown in Fig.~\ref{fig:uav}, both UAVs have the options of either moving in four directions or staying at their original positions. They do not know exact positions of their own and others, but can receive the signals of relative positions from each other. Since the two UAVs act simultaneously, one UAV needs to have a good estimation of what the other behaves so as to achieve its own goal. In our experiments, we let agent $i$ act as a chaser UAV who is planning to intercept a fugitive UAV $j$ on its way to the safe house. The chaser agent $i$ gets rewarded once it captures the fugitive agent $j$ before the agent $j$ reaches the safe house. We build the I-DID models for the chaser agent $i$ who models the fugitive agent $j$'s behaviors through the IDID, IDID-MDP and IDID-MDF algorithms. 

\begin{figure}[!ht]
\centering
\includegraphics[width=6cm]{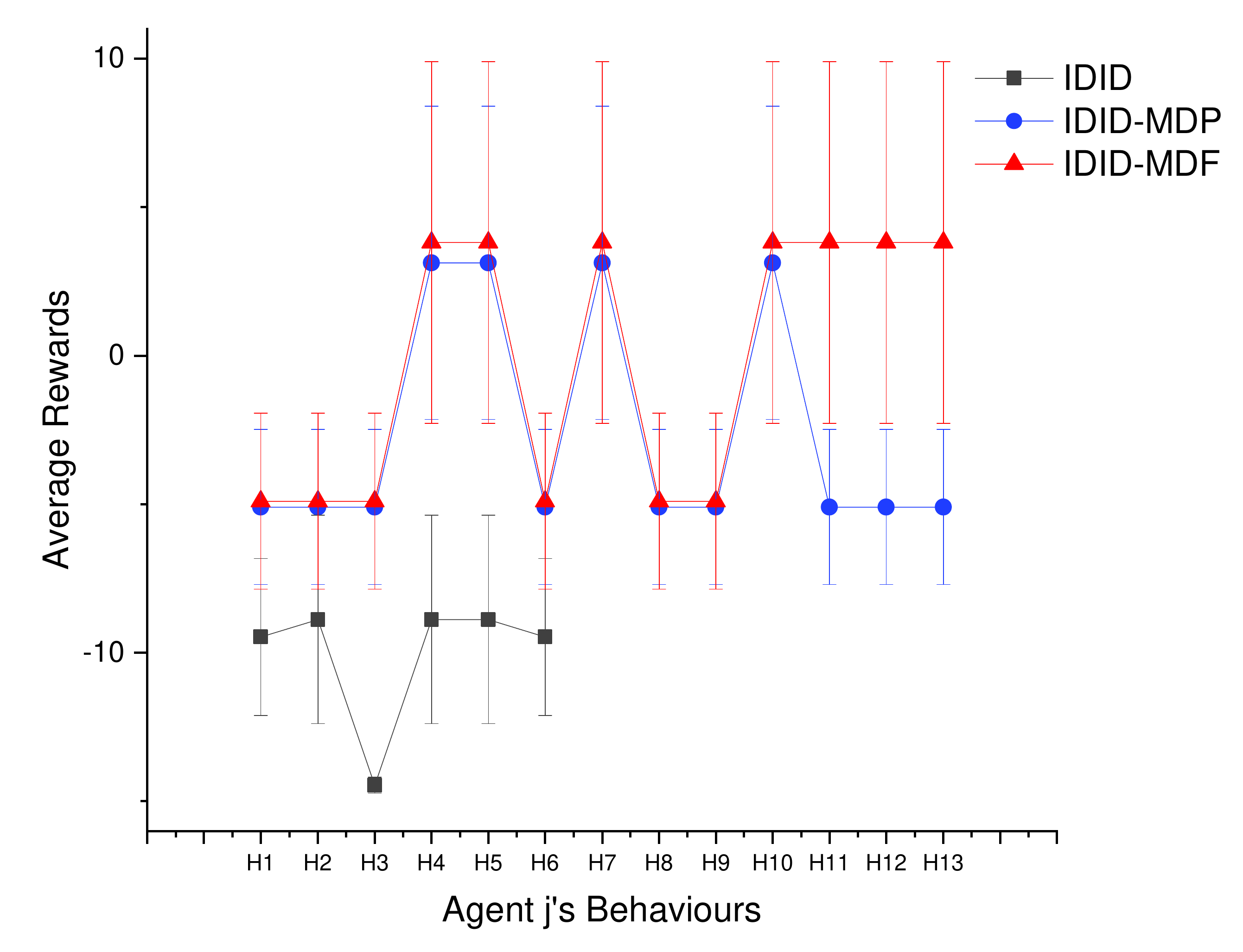}
\includegraphics[width=6cm]{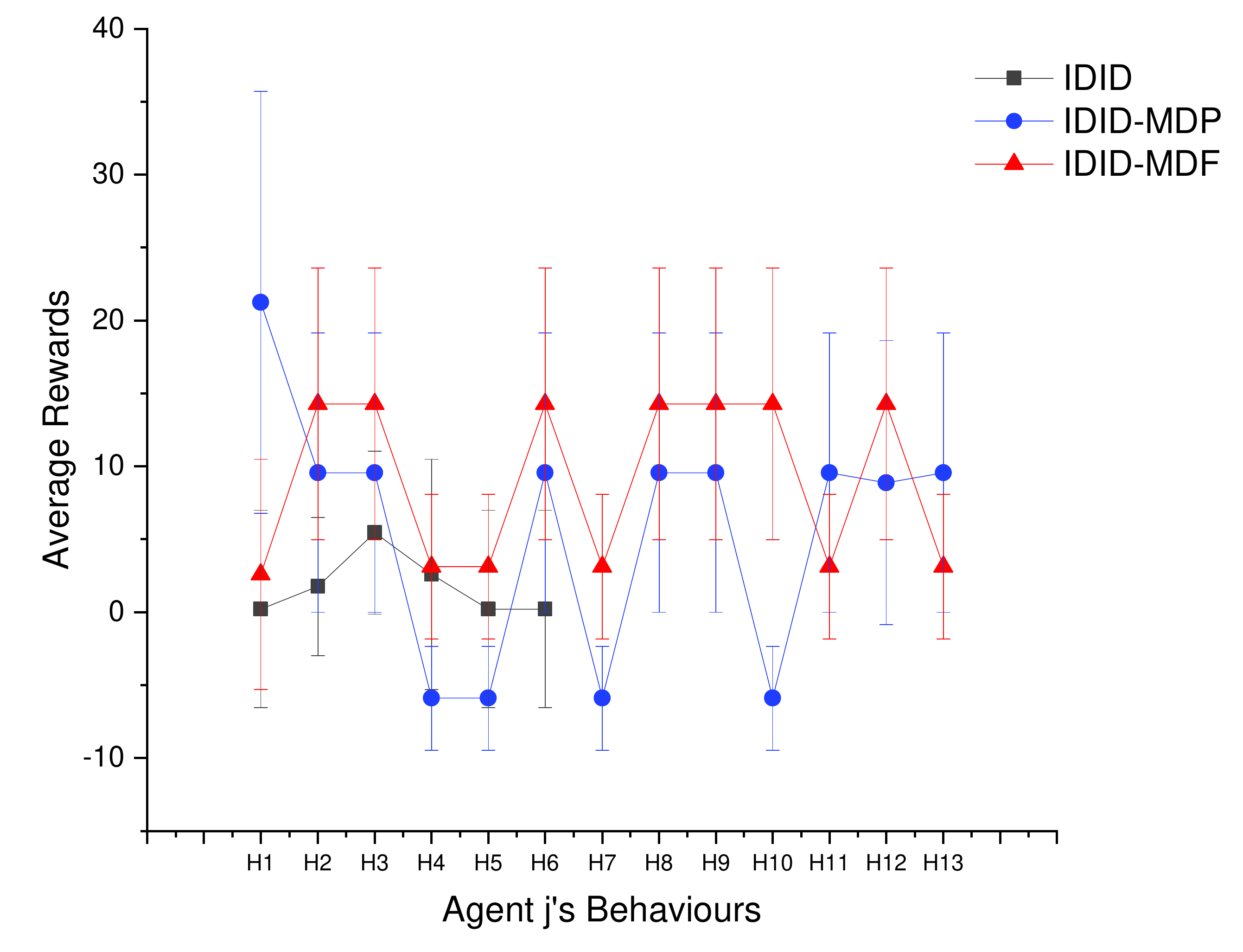}
\hspace{3cm}($a$)$T$=3\hspace{5cm}($b$)$T$=4
\caption{Average rewards received by the chaser agent $i$ when it plans to capture the fugitive agent $j$ through solving the I-DID models of ($a$) $T$=3 and ($b$) $T$=4.}
\label{fig:offlineUAV}
\end{figure}

In Fig.~\ref{fig:offlineUAV}, we report the average rewards of agent $i$ in the setting of $M$=6 for both the I-DID models of $T$=3 and $T$=4. The IDID-MDF algorithm exhibits better performance, compared to the IDID and IDID-MDP algorithms, similarly as shown in the multiagent problem domain. In most of the cases, both the IDID-MDP and IDID-MDF algorithms outperform the IDID algorithm particularly in the I-DID models with $T$=3 and $T$=4. Surprisingly, we notice that  the IDID algorithm does not perform good particularly in the I-DID models of $T$=3. For a short planning horizon, the chaser agent $i$ does not have 
enough times  to gather sufficient information as so to reduce the uncertainty of agent $j$'s behaviors, therefore failing in capturing the fugitive agent $j$ in most the cases. With extra models introduced by IDID-MDP and IDID-MDF, the chaser agent $i$ gains more knowledge about the fugitive agent $j$'s behaviors and achieves better rewards. The IDID-MDF algorithm performs slightly better than the IDID-MDP algorithm in the I-DID models of $T$=4. Capturing general patterns~(in IDID-MDF) provides more informative behaviors when the fugitive agent $j$ plans a long way to the safe house.

\begin{figure}[!ht]
\centering
\includegraphics[width=6cm]{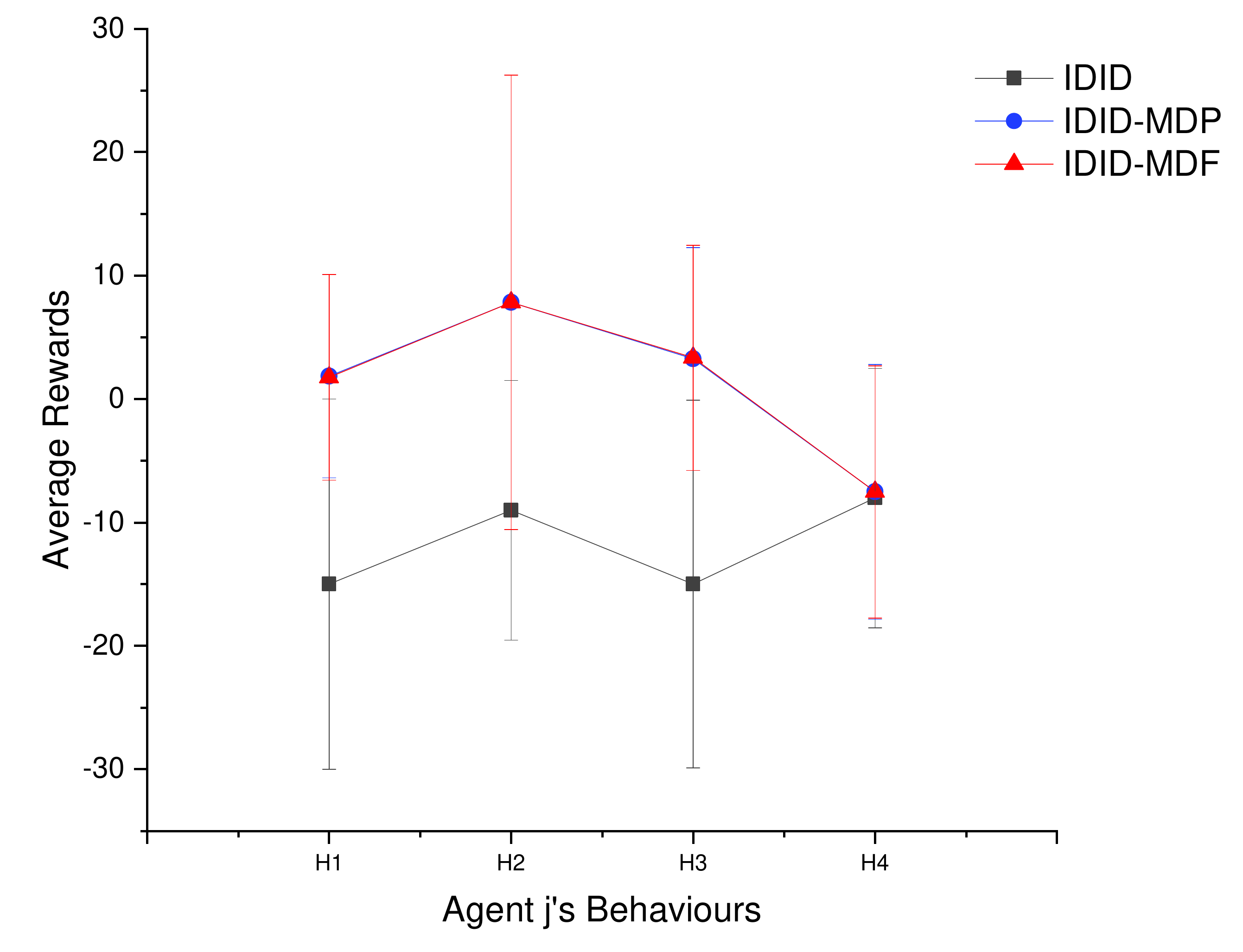}
\includegraphics[width=6cm]{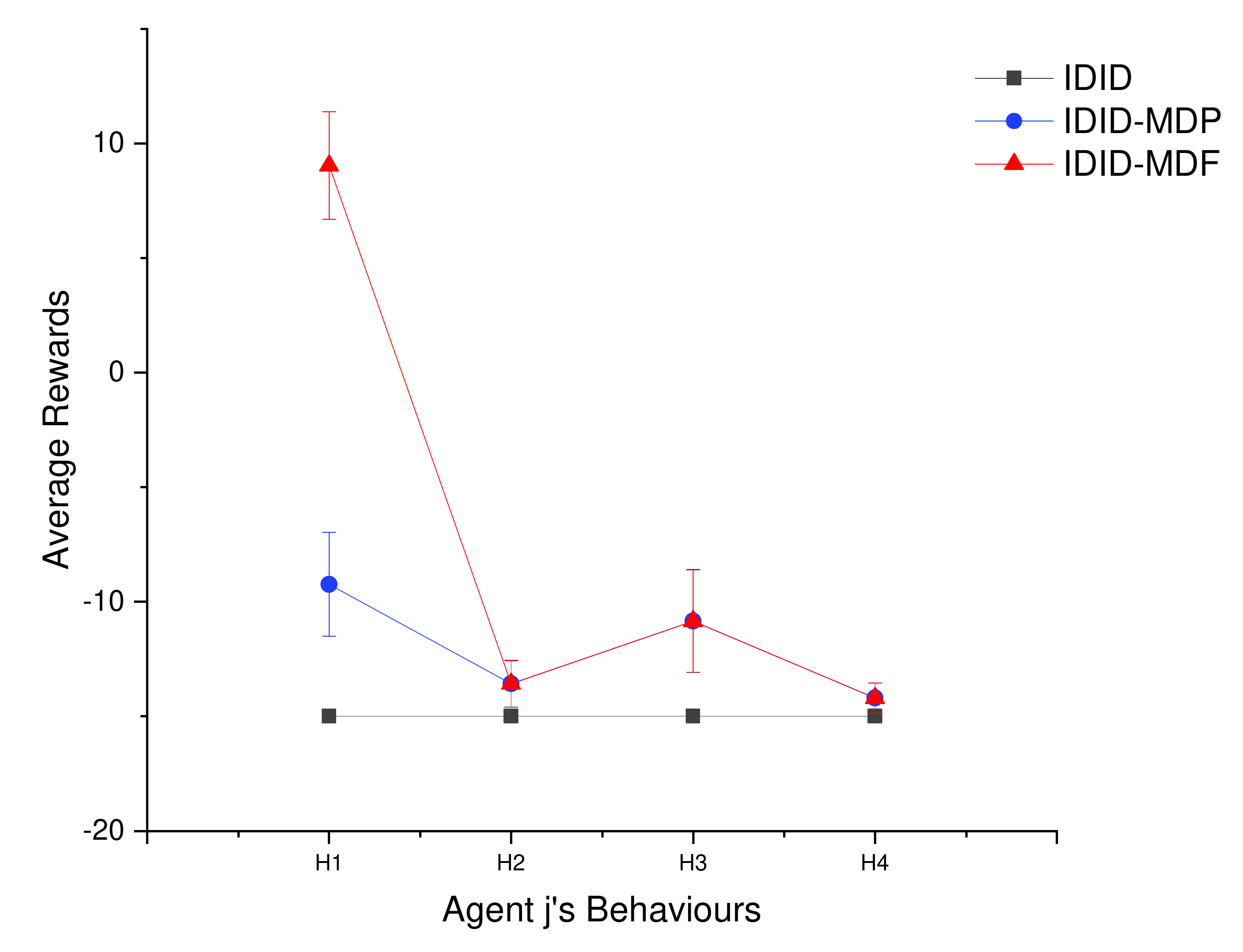}
\hspace{3cm}($a$)$T$=3\hspace{5cm}($b$)$T$=4
\caption{Average rewards received by the chaser agent $i$ when it knows the true model of the fugitive agent $j$ within the candidate model set in the I-DID models with ($a$)~$T$=3 and ($b$)~$T$=4.}
\label{fig:originuav}
\end{figure}

\begin{figure}[!ht]
\centering
\includegraphics[width=6cm]{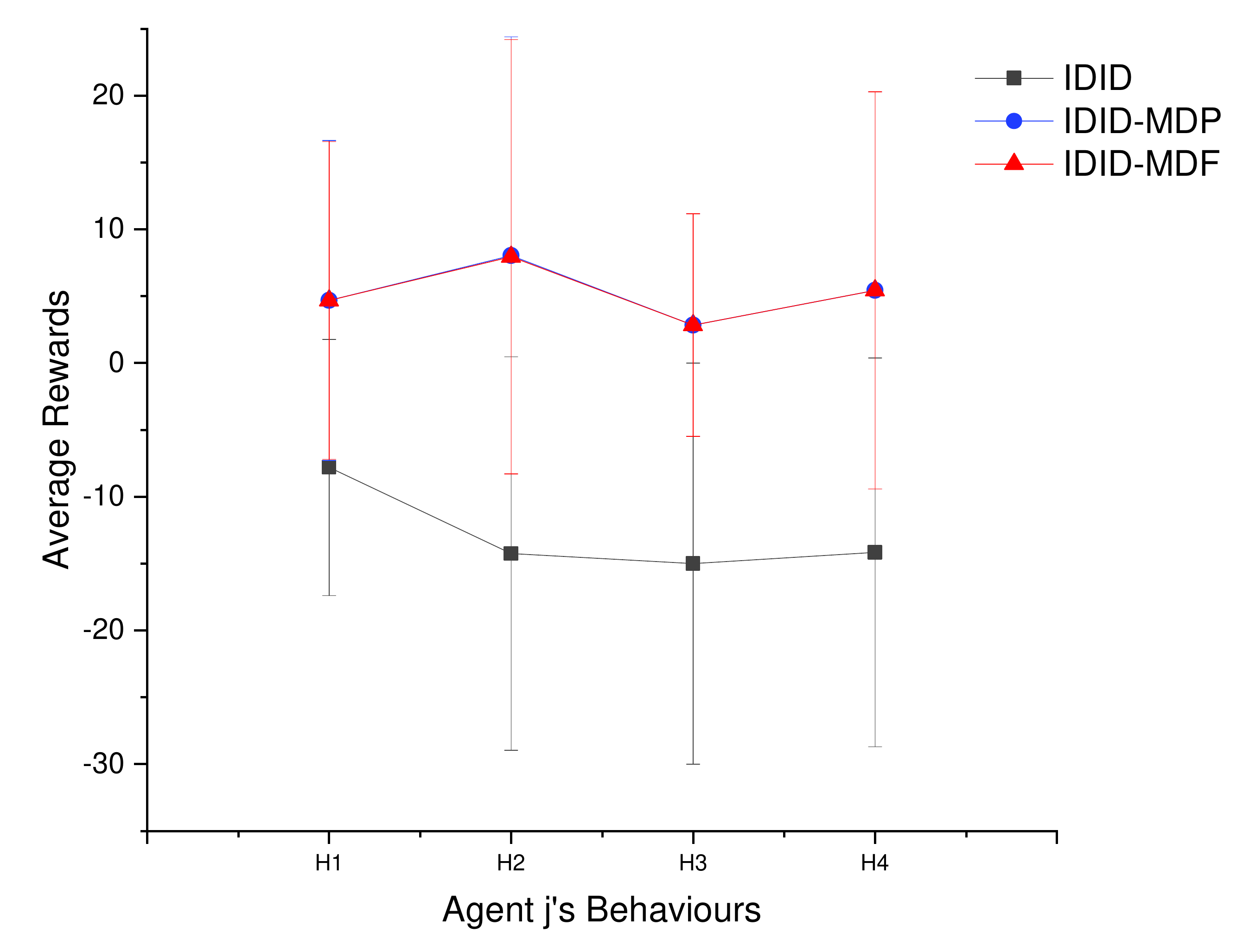}
\includegraphics[width=6cm]{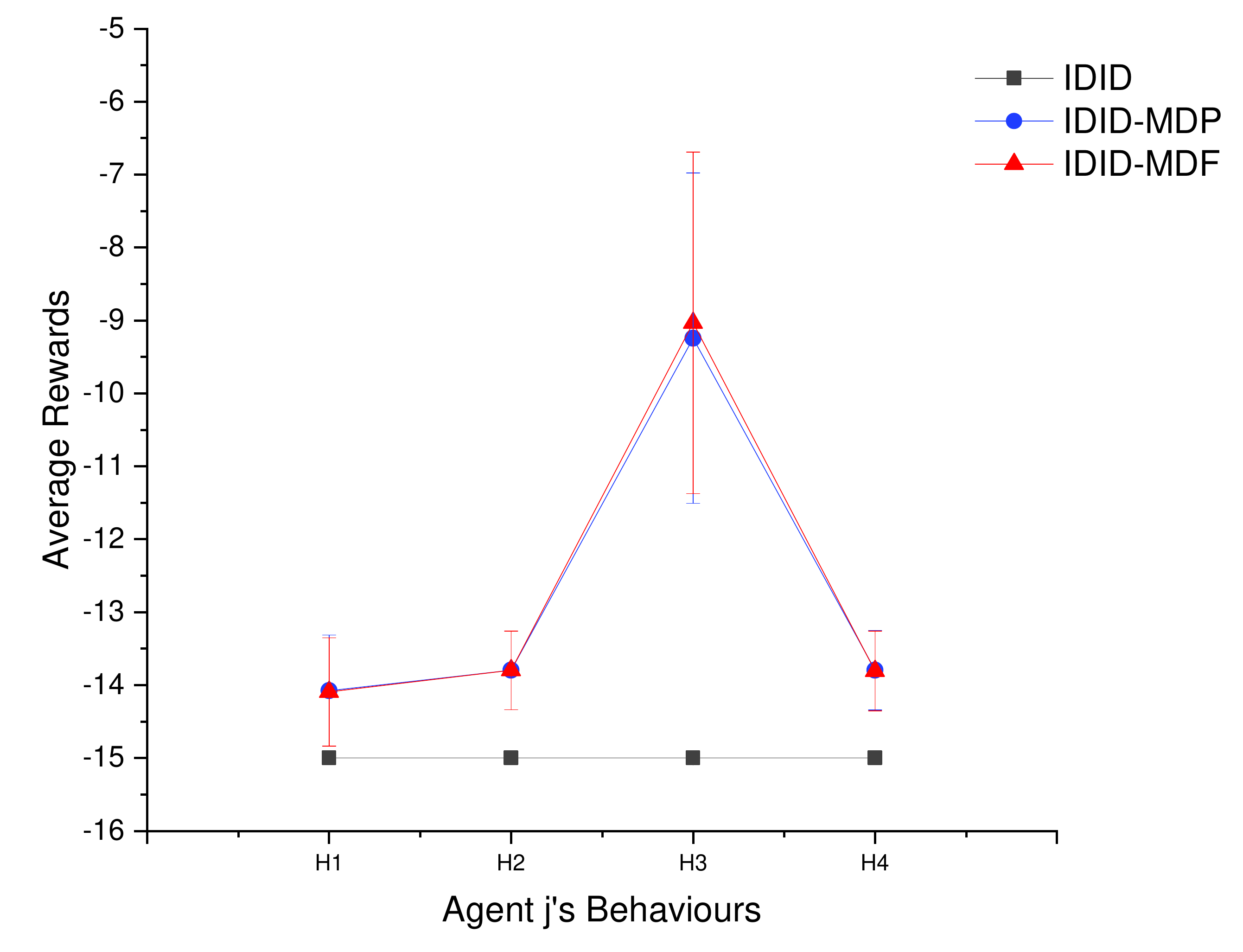}
\hspace{3cm}($a$)$T$=3\hspace{5cm}($b$)$T$=4
\caption{Average rewards received by the chaser agent $i$ when it faces random behaviors of the fugitive agent $j$ in the I-DID models ($a$)~$T$=3 and ($b$)~$T$=4.}
\label{fig:randomuav}
\end{figure}

We proceed to examine the performance of the IDID, IDID-MDP and IDID-MDF algorithms when ($a$) the chaser agent $i$ knows the true model of the fugitive agent $j$ in Fig.~\ref{fig:originuav} and ($b$) the chaser agent $i$ interacts with the fugitive agent $j$ whose behaviors are not in the chaser's mind in Fig.~\ref{fig:randomuav}. Both the IDID-MDP and IDID-MDF algorithms perform better than IDID, which is consistently with what we observed in the multiagent tiger problem domain. The IDID-MDF algorithm performs similarly to the IDID-MDP algorithm in this problem domain. Since the fugitive agent $j$ has many paths leading to the safe house, the limited number of models do not provide general patterns, which compromises the benefits of IDID-MDF. When we increase the number of models, as shown in Fig.~\ref{fig:offlineUAV}, the IDID-MDF algorithm does show better performance. Hence, the IDID-MDF algorithm is suggested to be well used in complex behaviors of other agents.

\begin{figure}[!ht]
\centering
\includegraphics[width=7cm]{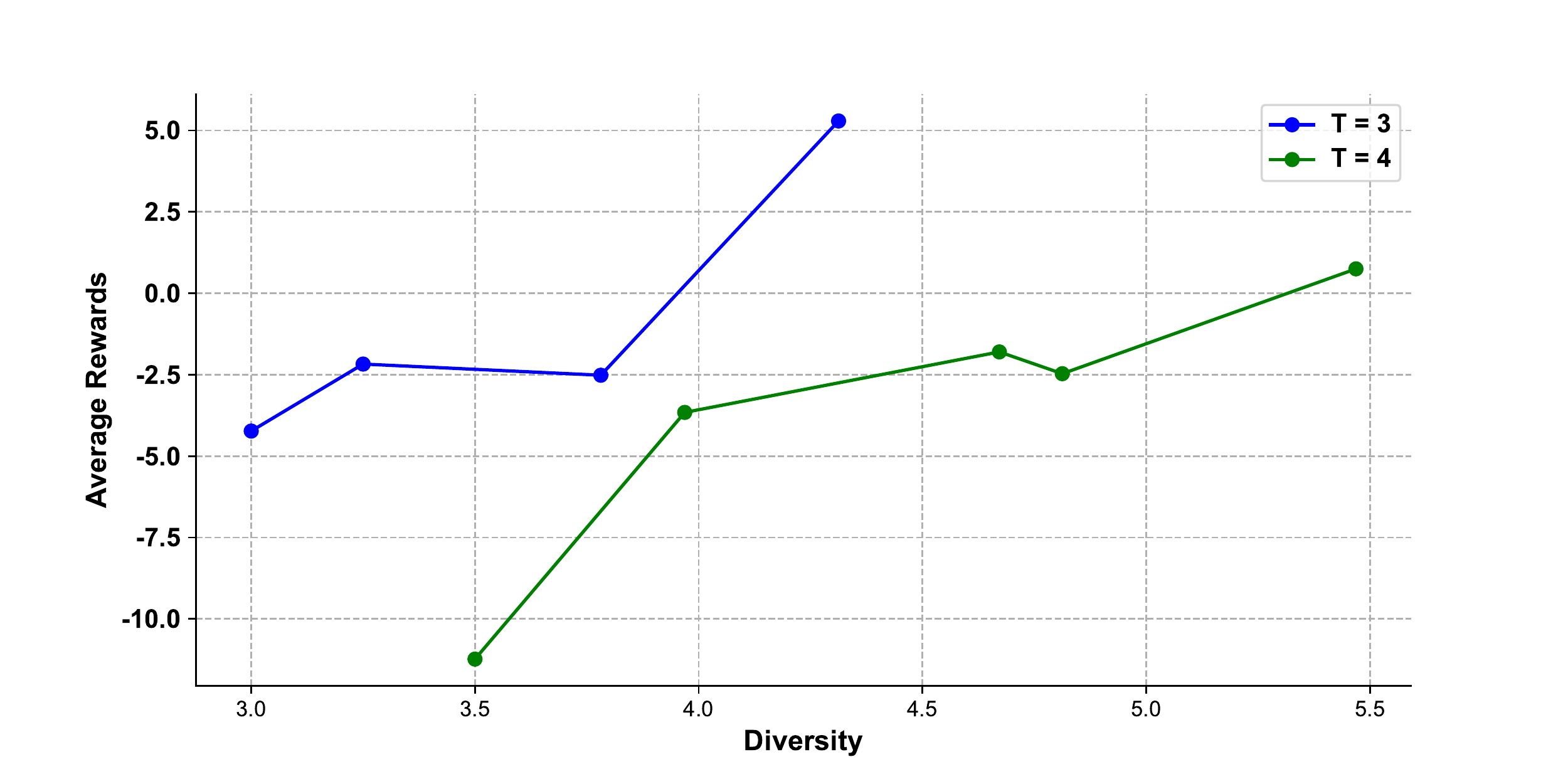}
\includegraphics[width=7cm]{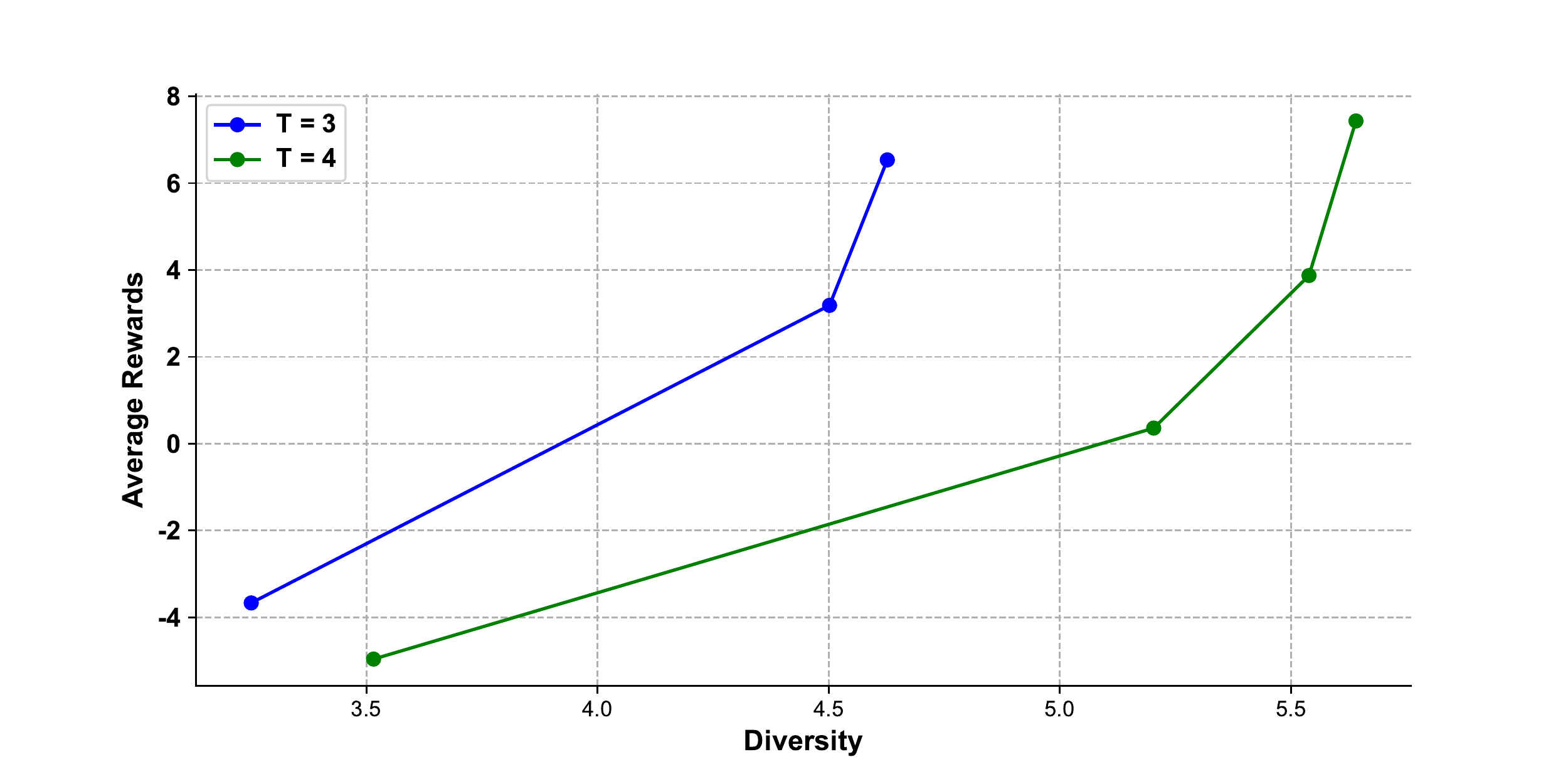}
\hspace{4cm}($a$) MDP Performance \hspace{4cm}($b$) MDF Performance
\caption{The chaser agent $i$ gets better rewards when the diversity of the fugitive agent $j$'s models~(measured by ($a$) MDP and ($b$) MDF) is increased in the I-DID models with $T$=3 and $T$=4.}
\label{fig:diversityuav}
\end{figure}

Additionally, we show the increasing rewards for the chaser agent $i$ when it holds a more diverse set of candidate models for the fugitive agent $j$ in Fig.~\ref{fig:diversityuav}. We observe that MDP has a small dip of the average rewards when the diversity is increased. In contrast, the rewards increase monotonically with the increasing values of the MDF diversity.
 We notice that the factor of capturing general behavior patterns does not contribute much into the MDF measurement, as shown the diversity values in the $x$-axis of the figure. This indicates that both the IDID-MDP and IDID-MDF algorithms may select the same models for the fugitive agent $j$. Thus, the two algorithms provide similar performance in the aforementioned experiments.

In summary, both the IDID-MDF and IDID-MDP algorithms perform as we expect in the experiments, and the IDID-MDF algorithm would be a better choice when more complex behaviors are involved in a problem domain.
 Furthermore, we show the comparative efficiency for the IDID-MDP and IDID-MDF algorithms in Table~\ref{tbl:times}.   For every set of initial models~($M$) of agent $j$, we compare their running times including generating new models and then selecting the top-$K$ models from the models in the Alg.~\ref{algorithm:topk}. Since MDF considers the extra factor of behavioral patterns in selecting the models, it can differentiate more behaviors therefore resulting in more new models in the model generation.  Apparently, MDF needs more times in comparing more models. However, its efficiency is not significantly compromised as shown in the table - not more than double amount of times as spent by MDP in most the experiments.

\begin{table}[!htbp]
  \centering
  \caption{Running times spent by the MDP and MDF methods on generating and selecting the top-$K$ models in Alg.~\ref{algorithm:topk}.}
    \begin{tabular}{|c|l|l|l|l|l|l|l|l|}
    \toprule
\cmidrule{2-9}          & \multicolumn{2}{c|}{Tiger} & \multicolumn{2}{c|}{UAV} & \multicolumn{2}{c|}{Tiger} & \multicolumn{2}{c|}{UAV} \\
    \midrule
    $M$     & \multicolumn{1}{c|}{MDP} & \multicolumn{1}{c|}{MDF} & \multicolumn{1}{c|}{MDP} & \multicolumn{1}{c|}{MDF} & \multicolumn{1}{c|}{MDP} & \multicolumn{1}{c|}{MDF} & \multicolumn{1}{c|}{MDP} & \multicolumn{1}{c|}{MDF} \\
    \midrule
    3     & 0.35  & 0.97  & 2.37  & 6.86  & 0.41  & 1.80  & 2.75  & 6.96  \\
    \midrule
    4     & 1.66  & 1.65  & 2.94  & 9.26  & 1.79  & 2.25  & 4.02  & 9.52  \\
    \midrule
    5     & 1.98  & 4.49  & 10.43  & 10.54  & 2.14  & 11.29  & 24.20  & 10.42  \\
    \midrule
    6     & 41.56  & 76.32  & 17.14  & 10.46  & 45.16  & 100.74  & 18.92  & 22.30  \\
    \midrule
    7     & 328.43  & 442.94  & 42.76  & 81.96  & 330.31  & 573.85  & 46.46  & 97.40  \\
    \bottomrule
    \end{tabular}%
  \label{tbl:times}%
\end{table}%

\section{Conclusion}
\label{sec:summary}
This work is the first attempt on studying the behavior diversity in I-DID solutions, which is significantly different from the previous I-DID research on compressing other agents' models in I-DIDs. Given known behaviors of other agents, we use a linear reduction technique to generate a set of representative behavior sequences. The resulting sequences are linearly independent and provide basic knowledge to generating new behaviors for other agents. Subsequently, we  sample a new set of behaviors, represented by policy trees, through  applying conventional inference techniques in the I-DID model. We propose two diversity measurements to select top-$K$ policy trees that maximise the diversity of the selected behaviors. We conduct experiments to investigate the performance of the new I-DID solutions on dealing with unknown behaviors of other agents in two problem domains.

As observed in the experimental study, it seems that the two diversity measurements are not one-size-fits-all. However, the study lights up critical thoughts about behavior diversity and its impact on a subject agent's decision quality. Thoughts are also captured by the recent I-DID study although the work still focuses on known behaviors of other agents~\cite{PanMTZ22}. Hence, this work will open a new field on developing I-DID solutions based on diversifying other agents' behaviors. A couple of new research lines would be conducted in the future. For example, a new method of diversifying behaviors with the consideration of a subject agent's decision quality would be immediate improvement of the current measurements. Of course, the challenge lies in measuring the decision quality since the subject agent has not yet completed the modelling process at this stage. Another way of developing new diversity measurements could be conducted in an online manner  when the subject agent receives more knowledge about true behaviors of other agents in real-time interactions. The diversity needs to be adjusted in a dynamic way when a specific type of behavior is discovered in the interactions. 
The aforementioned work can also benefit the recent research on intelligent systems when the systems are to be deployed in an open world and can not fully model a new environment prior to their applications.

\section*{Acknowledgements}
Professor Yifeng Zeng received the EPSRC New Investigator Award~(Grant No. EP/S011609/1) and Dr. Biyang Ma conducted the research under the EPSRC project.
This work is supported in part by the National Natural Science Foundation of China (Grants No.62176225, 61772442 and 61836005). 




\bibliography{sample}

\begin{thebibliography}{28}
\providecommand{\natexlab}[1]{#1}
\providecommand{\url}[1]{\texttt{#1}}
\providecommand{\urlprefix}{}

\bibitem[{Poh et~al.(1994)Kim Leng Poh and Michael R. Fehling and Eric Joel
  Horvitz}]{PohFH94}
Poh KL, Fehling MR, Horvitz EJ.
\newblock Dynamic Construction and Refinement of Utility-Based Categorization
  Models.
\newblock {IEEE} Trans Syst Man Cybern Syst 1994;24(11):1653--1663.

\bibitem[{Albrecht and Stone(2018)Stefano V. Albrecht and Peter
  Stone}]{AlbrechtS18}
Albrecht SV, Stone P.
\newblock Autonomous agents modelling other agents: {A} comprehensive survey
  and open problems.
\newblock Artif Intell 2018;258:66--95.

\bibitem[{Bernstein et~al.(2002)Bernstein, Daniel and Givan, Robert and
  Immerman, Neil and Zilberstein, Shlomo}]{Bernstein02}
Bernstein D, Givan R, Immerman N, Zilberstein S.
\newblock The Complexity of Decentralized Control of Markov Decision Processes.
\newblock Mathematics of Operations Research 2002 12;27:637--842.

\bibitem[{Zeng and Doshi(2012)Yifeng Zeng and Prashant Doshi}]{ZengD12}
Zeng Y, Doshi P.
\newblock Exploiting Model Equivalences for Solving Interactive Dynamic
  Influence Diagrams.
\newblock J Artif Intell Res 2012;43:211--255.

\bibitem[{Doshi et~al.(2020)Prashant Doshi and Piotr J. Gmytrasiewicz and
  Edmund H. Durfee}]{DoshiGD20}
Doshi P, Gmytrasiewicz PJ, Durfee EH.
\newblock Recursively modeling other agents for decision making: {A} research
  perspective.
\newblock Artif Intell 2020;279.

\bibitem[{Doshi et~al.(2009)Prashant Doshi and Yifeng Zeng and Qiongyu
  Chen}]{DoshiZC09}
Doshi P, Zeng Y, Chen Q.
\newblock Graphical models for interactive POMDPs: representations and
  solutions.
\newblock Auton Agents Multi Agent Syst 2009;18(3):376--416.

\bibitem[{{Tatman} and {Shachter}(1990)J. A. {Tatman} and R. D.
  {Shachter}}]{Tatman90}
{Tatman} JA, {Shachter} RD.
\newblock Dynamic programming and influence diagrams.
\newblock IEEE Transactions on Systems, Man, and Cybernetics
  1990;20(2):365--379.

\bibitem[{S{\o}ndberg{-}Jeppesen et~al.(2013)Nicolaj S{\o}ndberg{-}Jeppesen and
  Finn Verner Jensen and Yifeng Zeng}]{Sondberg-JeppesenJZ13}
S{\o}ndberg{-}Jeppesen N, Jensen FV, Zeng Y.
\newblock Opponent modeling in a {PGM} framework.
\newblock In: Proceedings of International conference on Autonomous Agents and
  Multi-Agent Systems~(AAMAS); 2013. p. 1149--1150.

\bibitem[{Conroy et~al.(2016)Ross Conroy and Yifeng Zeng and Jing
  Tang}]{ConroyZT16}
Conroy R, Zeng Y, Tang J.
\newblock Approximating Value Equivalence in Interactive Dynamic Influence
  Diagrams Using Behavioral Coverage.
\newblock In: Kambhampati S, editor. Proceedings of the Twenty-Fifth
  International Joint Conference on Artificial Intelligence~(IJCAI); 2016. p.
  201--207.

\bibitem[{Pan et~al.(2022)Yinghui Pan and Biyang Ma and Jing Tang and Yifeng
  Zeng}]{PanMTZ22}
Pan Y, Ma B, Tang J, Zeng Y.
\newblock Behavioral model summarisation for other agents under uncertainty.
\newblock Information Sciences 2022;582:495--508.

\bibitem[{Andersen et~al.(2020)Andersen, P.A. and Goodwin, M. and Granmo,
  O.C.}]{rl}
Andersen PA, Goodwin M, Granmo OC.
\newblock Towards safe reinforcement-learning in industrial grid-warehousing.
\newblock Information Sciences 2020;537:467 -- 484.

\bibitem[{Albrecht et~al.(2020)Stefano V. Albrecht and Peter Stone and Michael
  P. Wellman}]{Stefano20}
Albrecht SV, Stone P, Wellman MP.
\newblock Special issue on autonomous agents modelling other agents: Guest
  editorial.
\newblock Artificial Intelligence 2020;285:103292.

\bibitem[{Brown(1951)George W. Brown}]{brown:fp1951}
Brown GW.
\newblock Iterative Solution of Games by Fictitious Play.
\newblock New York, USA: Wiley; 1951.

\bibitem[{Alessandro and Piotr(2017)Panella Alessandro and Gmytrasiewicz
  Piotr}]{Panella:2017}
Alessandro P, Piotr G.
\newblock Interactive POMDPs with Finite-state Models of Other Agents.
\newblock Autonomous Agents and Multi-Agent Systems 2017 Jul;31(4):861--904.

\bibitem[{Gmytrasiewicz and Doshi(2005)Piotr Gmytrasiewicz and Prashant
  Doshi}]{Gmytrasiewicz05:Framework:JAIR}
Gmytrasiewicz P, Doshi P.
\newblock A Framework for Sequential Planning in Multiagent Settings.
\newblock Journal of Artificial Intelligence Research (JAIR) 2005;24:49--79.

\bibitem[{Bolander and Andersen(2011)Thomas Bolander and Mikkel Birkegaard
  Andersen}]{Thomas:DEL}
Bolander T, Andersen MB.
\newblock Epistemic planning for single- and multi-agent systems.
\newblock Journal of Applied Non-Classical Logics 2011;21(1):9--33.

\bibitem[{Ma et~al.(2021)Yuxi Ma and Meng Shen and Yuhang Zhao and Zhao Li and
  Xiaoyao Tong and Quanxin Zhang and Zhi Wang}]{ma21}
Ma Y, Shen M, Zhao Y, Li Z, Tong X, Zhang Q, et~al.
\newblock Opponent portrait for multiagent reinforcement learning in
  competitive environment.
\newblock International Journal of Intelligent Systems 2021;36:7461‐7474.

\bibitem[{Suryadi and Gmytrasiewicz(1999)Suryadi, D. and Gmytrasiewicz, P.
  J}]{Dicky99:ID}
Suryadi D, Gmytrasiewicz PJ.
\newblock Learning models of other agents using influence diagrams.
\newblock In: International Conference on User Modeling Springer; 1999. p.
  223--232.

\bibitem[{Koller and Milch(2003)Koller, Daphne and Milch, Brian}]{maid}
Koller D, Milch B.
\newblock Multi-agent influence diagrams for representing and solving games.
\newblock Games and Economic Behavior 2003;45(1):181--221.

\bibitem[{Gal and Pfeffer(2008)Gal, Ya'akov and Pfeffer, Avi}]{nid}
Gal Y, Pfeffer A.
\newblock Networks of Influence Diagrams: A Formalism for Representing Agents'
  Beliefs and Decision-Making Processes.
\newblock Journal of Artificial Intelligence Research 2008 sep;33(1):109–147.

\bibitem[{Pynadath and Marsella(2007)David V. Pynadath and Stacy
  Marsella}]{PynadathM07}
Pynadath DV, Marsella S.
\newblock Minimal Mental Models.
\newblock In: Proceedings of the Twenty-Second {AAAI} Conference on Artificial
  Intelligence; 2007. p. 1038--1044.

\bibitem[{Rathnasabapathy et~al.(2006)Bharaneedharan Rathnasabapathy and
  Prashant Doshi and Piotr J. Gmytrasiewicz}]{Bha06}
Rathnasabapathy B, Doshi P, Gmytrasiewicz PJ.
\newblock Exact solutions of interactive POMDPs using behavioral equivalence.
\newblock In: The Fifth International Joint Conference on Autonomous Agents and
  Multiagent Systems {ACM}; 2006. p. 1025--1032.

\bibitem[{Zeng et~al.(2016)Yifeng Zeng and Prashant Doshi and Yingke Chen and
  Yinghui Pan and Hua Mao and Muthukumaran Chandrasekaran}]{ZengDCPMC16}
Zeng Y, Doshi P, Chen Y, Pan Y, Mao H, Chandrasekaran M.
\newblock Approximating behavioral equivalence for scaling solutions of I-DIDs.
\newblock Knowledge Information Systems 2016;49(2):511--552.

\bibitem[{Conroy et~al.(2016)Conroy, R. and Zeng, Y. and Cavazza, M. and Tang,
  J. and Pan, Y.}]{Ross16}
Conroy R, Zeng Y, Cavazza M, Tang J, Pan Y.
\newblock A Value Equivalence Approach for Solving Interactive Dynamic
  Influence Diagrams.
\newblock In: Proceedings of the 15th International Conference on Autonomous
  Agents{\&} Multiagent Systems (AAMAS); 2016. p. 1162--1170.

\bibitem[{Pan et~al.(2021)Yinghui Pan and Jing Tang and Biyang Ma and Yifeng
  Zeng and Zhong Ming}]{Pan0MZ021}
Pan Y, Tang J, Ma B, Zeng Y, Ming Z.
\newblock Toward data-driven solutions to interactive dynamic influence
  diagrams.
\newblock Knowl Inf Syst 2021;63(9):2431--2453.

\bibitem[{Jensen and Nielsen(2007)Jensen, Finn V. and Nielsen, Thomas D.}]{pgm}
Jensen FV, Nielsen TD.
\newblock Bayesian Networks and Decision Graphs.
\newblock 2nd ed. Springer; 2007.

\bibitem[{Doshi et~al.(2009)Prashant Doshi and Yifeng Zeng and Qiongyu
  Chen}]{Doshi09:Graphical}
Doshi P, Zeng Y, Chen Q.
\newblock Graphical Models for Interactive POMDPs: Representations and
  Solutions.
\newblock Journal of Autonomous Agents and Multi-Agent Systems (JAAMAS)
  2009;18(3):376--416.

\bibitem[{Cunningham and Ghahramani(2015)John P. Cunningham and Zoubin
  Ghahramani}]{JMLR16}
Cunningham JP, Ghahramani Z.
\newblock Linear Dimensionality Reduction: Survey, Insights, and
  Generalizations.
\newblock Journal of Machine Learning Research 2015;16(89):2859--2900.

\end{thebibliography}



\end{document}